\documentclass{article}

\usepackage{microtype}
\usepackage{graphicx}
\usepackage{subfigure}
\usepackage{booktabs} 

\usepackage{hyperref}

\usepackage[accepted]{icml2020}

\usepackage{hyperref}
\usepackage{url}
\usepackage{multirow}
\usepackage{amsmath, bm}
\usepackage{amsfonts,amsthm}
\usepackage{wrapfig}
\usepackage{xfrac}
\usepackage{sidecap}
\usepackage{adjustbox}
\usepackage{caption}

\newcommand{\var}{{\rm I\kern-.3em D}}

\newcommand{\cond}{\,|\,}
\newcommand{\Normal}{\mathcal{N}}

\newcommand{\eps}{\varepsilon}

\newtheorem*{theorem*}{Theorem}

\newtheorem{proposition}{Proposition}
\newtheorem{trick}{Trick}

\icmltitlerunning{Involutive MCMC}

\begin{document}

\twocolumn[
\icmltitle{Involutive MCMC: a Unifying Framework}



\begin{icmlauthorlist}
\icmlauthor{Kirill Neklyudov}{samsung,samsunghse}
\icmlauthor{Max Welling}{uva,cifar}
\icmlauthor{Evgenii Egorov}{skoltech}
\icmlauthor{Dmitry Vetrov}{samsung,samsunghse}
\end{icmlauthorlist}

\icmlaffiliation{samsung}{Samsung AI Center Moscow}
\icmlaffiliation{samsunghse}{Samsung-HSE Laboratory, National Research University Higher School of Economics}
\icmlaffiliation{uva}{University of Amsterdam}
\icmlaffiliation{cifar}{Canadian Institute for Advanced Research}
\icmlaffiliation{skoltech}{Skolkovo Institute of Science and Technology}

\icmlcorrespondingauthor{Kirill Neklyudov}{k.necludov@gmail.com}

\icmlkeywords{Machine Learning, ICML, MCMC}

\vskip 0.3in
]



\printAffiliationsAndNotice{}  

\begin{abstract}
Markov Chain Monte Carlo (MCMC) is a computational approach to fundamental problems such as inference, integration, optimization, and simulation. The field has developed a broad spectrum of algorithms, varying in the way they are motivated, the way they are applied and how efficiently they sample. 
Despite all the differences, many of them share the same core principle, which we unify as the Involutive MCMC (iMCMC) framework. Building upon this, we describe a wide range of MCMC algorithms in terms of iMCMC, and formulate a number of ``tricks'' which one can use as design principles for developing new MCMC algorithms. Thus, iMCMC provides a unified view of many known MCMC algorithms, which facilitates the derivation of powerful extensions. We demonstrate the latter with two examples where we transform known reversible MCMC algorithms into more efficient irreversible ones.
\end{abstract}

\section{Introduction}

Machine learning algorithms with stochastic latent variables or parameters (a.k.a. Bayesian models) require often intractable posterior inference over these unobserved random variables. The most popular approach these days are variational approximations where possibly complex and/or amortized posterior distributions are optimized and used for inference: e.g. $q(z|x)$ for latent variables or $q(\theta)$ for parameters. However, these distributions are usually biased and may not be easy to optimize. 

\begin{table}[]
    \centering
    \begin{tabular}{c|c}
        Name \& Citation & Appendix \\
        \hline
        Metropolis-Hastings \citep{hastings1970monte} & \ref{app:MH}\\
        Mixture Proposal \citep{habib2018auxiliary} & \ref{app:MPMCMC} \\
        Multiple-Try Metropolis \citep{liu2000multiple} & \ref{app:MTM} \\
        Sample-Adaptive MCMC \citep{zhu2019sample} & \ref{app:SAMCMC} \\
        Reversible-Jump MCMC \citep{green1995reversible} & \ref{app:RJMCMC}\\
        Hybrid Monte Carlo \citep{duane1987hybrid} & \ref{app:HMC}\\
        RMHMC \citep{girolami2011riemann} & \ref{app:RMHMC}\\
        NeuTra \citep{hoffman2019neutra} & \ref{app:neutra}\\
        A-NICE-MC \citep{song2017nice} & \ref{app:nicemc}\\
        L2HMC \citep{levy2017generalizing} & \ref{app:l2hmc}\\
        Persistent HMC \citep{horowitz1991generalized} & \ref{app:horowitz}\\
        Gibbs \citep{geman1984stochastic} & \ref{app:gibbs}\\
        Look Ahead \citep{sohl2014hamiltonian} & \ref{app:LAHMC} \\
        NRJ \citep{gagnon2019non} & \ref{app:nrj}\\
        Lifted MH \citep{turitsyn2011irreversible} & \ref{app:lmh}\\
    \end{tabular}
    \caption{List of algorithms that we describe by the Involutive MCMC framework.
    See their descriptions and formulations in terms of iMCMC in corresponding appendices.}
    \label{tab:generalization_list}
\vspace{-1em}
\end{table}

A completely different class of algorithms is given by MCMC algorithms. Here we design a stochastic process that eventually samples from the correct (i.e. target) distribution. This has the advantage that we are guaranteed to obtain unbiased samples at the cost of possibly slow mixing and long burn-in times. There is a huge literature on MCMC algorithms across many different scientific fields such as statistics, bio-informatics, physics, chemistry, machine learning etc. 

More recently, researchers have started to design MCMC kernels by using learnable components, in particular flows which are also often used in variational approaches \citep{song2017nice, hoffman2019neutra}. We anticipate these hybrid approaches will become an important family of inference methods for approximate inference. 

In this paper we provide a unifying framework for MCMC algorithms, including the hybrid approaches mentioned above. We call this Involutive MCMC (iMCMC). We provide an overview of many existing MCMC methods reformulated as iMCMC algorithms. See table \ref{tab:generalization_list} for the list. The power of our framework is the ease with which one can now start to combine and improve these algorithms using a number of ``tricks'' that we discuss extensively. We provide two examples for how this generalization might work in the experiments section. We hope our work might spur the development of new approximate inference methods based on ideas from MCMC inference.

We summarize the main contributions of the paper as follows.
\begin{itemize}
    \item In Section \ref{sec:iMCMC}, we introduce the Involutive MCMC formalism to describe a wide range of MCMC algorithms.
    This formalism provides a simple way to verify invariance of the target distribution, at the same time highlighting the main constraints this invariance put on the design of samplers.
    \item In Section \ref{sec:tricks}, we summarize the main ideas of different MCMC algorithms in the literature, providing the reader with the set of ``tricks''.
    These tricks provide a simple way to incorporate new features into a sampler in order to increase its efficiency, without re-deriving the fixed point equation or validating detailed balance.
    \item Finally, in Section \ref{sec:applications}, we demonstrate the potential of iMCMC formalism deriving irreversible counterparts of existent MCMC methods, and demonstrate empirical gains on different target distributions.
\end{itemize}

\section{Involutive MCMC}
\label{sec:iMCMC}

MCMC algorithms are designed by specifying a transition probability $t(x'\cond x)$ that maps a distribution $p_t$ to a new distribution $p_{t+1}$. Repeatedly applying this map to an initial distribution $p_0$ should result in the target distribution $p$. One can show that this is guaranteed if the map is ergodic (whose average over space is equal to its average over time, which is the number of applications of the map here) and leaves the target distribution invariant:
\begin{align}
    \int dx t(x'\cond x) p(x) = p(x').
    \label{eq:fpe}
\end{align}
Usually, one can not compute the full integral and so we approximate the process of iteratively applying the transition kernel by sampling a single sample from it at every iteration. At convergence, these samples will then be guaranteed to be distributed according to the target distribution. 
In the rest of the paper, we will refer to equation \eqref{eq:fpe} as \textit{the fixed point equation}.

The transition kernel is usually stochastic, but can also be deterministic, in which case it represents an iterated map. Applying it to a sample from $p_0$ will thus generate a deterministic trajectory. To be ergodic, this trajectory can not be periodic and is usually chaotic. Deterministic (irreversible) Markov chains can have very high mixing rates, which is the reason why we are interested in them. For a deterministic map we consider a transition kernel of the form $t(x'\cond x)= \delta(x'-f(x))$, where $f(x)$ is a bijection. Invariance then looks like:
\begin{equation}
    \int dx \delta(x'-f(x))p(x) = p(x').
\end{equation}
This equation immediately implies the measure-preserving condition
\begin{equation}
    p(x) = p(f(x)) \bigg| \frac{\partial f}{\partial x} \bigg| = p(f^{-1}(x)) \bigg| \frac{\partial f^{-1}}{\partial x} \bigg|,
\label{eq:measure_preserve}
\end{equation}
where $\frac{\partial f}{\partial x}$ denotes the Jacobian of $f(x)$.

If we find a map $f(x)$ that satisfies equation \eqref{eq:measure_preserve} and that will reach any point in the support of $p(x)$ through repeated application, we obtain a proper sampler (an ergodic chain with stationary distribution $p(x)$).
A practical example of such a sampler can be obtained analogously to \citep{murray2012driving, neal2012view} using the CDF of the target distribution and its inverse:
\begin{equation}
    f(x) = F^{-1}_{p}\bigg((F_{p}(x) + C) \text{ mod } 1\bigg),
    \label{eq:cdf_kernel}
\end{equation}
where $F_{p}$ denotes the CDF of the target density $p(x)$ and the constant $C$ can be chosen as an irrational number to guarantee ergodicity of the map.
To verify the correctness of this transition kernel one can straightforwardly put formula \eqref{eq:cdf_kernel} into equation \eqref{eq:measure_preserve} or treat it as a special case of algorithm by \cite{murray2012driving, neal2012view} (see Appendix \ref{app:det_mcmc}).

The equation \eqref{eq:measure_preserve} may be too restrictive, making the design of deterministic measure-preserving transformations (solutions of equation \eqref{eq:measure_preserve}) a very difficult task.
To the best of our knowledge, only the algorithm proposed in \citep{murray2012driving, neal2012view} provides practical examples of such transformations, relying on the knowledge of CDFs and their inverse.

In this paper, we propose a different transition kernel that leaves the target distribution invariant.
That is,
\begin{align}
\begin{split}
    t(x'\cond x) = \delta(x'-f(x)) \underbrace{\min\bigg\{1, \frac{p(f(x))}{p(x)} \bigg| \frac{\partial f}{\partial x}\bigg| \bigg\}}_{P_{\text{accept}}} + \\
    + \delta(x'-x) \underbrace{\bigg(1- \min\bigg\{1, \frac{p(f(x))}{p(x)} \bigg| \frac{\partial f}{\partial x}\bigg| \bigg\}\bigg)}_{P_{\text{reject}}},
    \label{eq:det_kernel}
\end{split}
\end{align}
Assume that $p_0$ is a delta-peak at some initial location $x_0$. Then the application of \ref{eq:det_kernel} will map this single delta peak to two delta peaks each with it's own weight: one peak at $x_0$ and the other at $f(x_0)$. At iteration $t$ there are thus $t$ weighted delta peaks, which becomes increasingly expensive to iterate forward. The more practical implementation of this kernel is to accept each new sample $x_{t+1}=f(x_t)$ with probability $P_{\text{accept}}$ (see equation \ref{eq:det_kernel}) or reject and keep the current point $x_t$ with probability $(1-P_{\text{accept}})$.

Putting this transition kernel into equation \eqref{eq:fpe}, we can simplify the fixed point equation to the condition that we formulate in the following proposition (see proof in Appendix \ref{app:fpe_condition}).
\begin{proposition}
\label{th:x_cond}
The fixed point equation \eqref{eq:fpe} for the transition kernel \eqref{eq:det_kernel} is equivalent to the equation
\begin{align}
\begin{split}
    \min\bigg\{p(f^{-1}(x))\bigg|\frac{\partial f^{-1}}{\partial x}\bigg| ,p(x)\bigg\} = \\ \min\bigg\{p(x), p(f(x)) \bigg| \frac{\partial f}{\partial x} \bigg| \bigg\}.
\label{eq:x_cond}
\end{split}
\end{align}
A similar equation can be derived for the Barker's acceptance test \citep{barker1965monte} (see Appendix \ref{app:fpe_condition}).
\end{proposition}

Firstly, we note that measure-preserving transformations (solutions of \eqref{eq:measure_preserve}) are a special case of solutions of \eqref{eq:x_cond}, with a zero rejection probability. Thus, these solutions eliminate all the stochasticity from the transition kernel \eqref{eq:det_kernel}, accepting all samples.
However, equation \eqref{eq:x_cond} accepts a broader family of solutions that can be described by the equation
\begin{align}
    p(f(x)) \bigg| \frac{\partial f}{\partial x} \bigg| = p(f^{-1}(x)) \bigg| \frac{\partial f^{-1}}{\partial x} \bigg|.
    \label{eq:involutive_cond}
\end{align}
The main difference with \eqref{eq:measure_preserve} is that here we do not restrict $f$ to preserve the target density.
Instead, we restrict $f(f(x))$ to preserve the target density:
\begin{align}
    p(x) = p(f(f(x))) \bigg| \frac{\partial f(f(x))}{\partial x} \bigg|.
\end{align}
The last equation can be obtained from equation \eqref{eq:involutive_cond} by considering the point $x=f(x')$.
At first glance, the problem of finding an $f$ such that $f(f(x))$ preserves the target measure is equally difficult to the problem of finding an $f(x)$ that preserves the density.
However, the class of functions called involutions solves eq. \ref{eq:involutive_cond} trivially because they satisfy $f(x) = f^{-1}(x)$.
In such a case, $f(f(x)) = x$ indeed preserves the target measure by being an identity mapping.
Thus, unlike equation \eqref{eq:measure_preserve}, equation \eqref{eq:x_cond} is solved by involutive functions $f$.
Unfortunately, it is not silver bullet: by inserting such $f$ into the transition kernel, eq. \eqref{eq:det_kernel} reduces our transition kernel to jump only between two points: from $x$ to $f(x)$ and then to $f(f(x)) = f^{-1}(f(x)) = x$ again.

To be able to cover the support of the target distribution with involutive $f$, we introduce an additional source of stochasticity into \eqref{eq:det_kernel}. We do this through an \textit{auxiliary variable}.
That is, instead of traversing the target $p(x)$, we traverse the distribution $p(x,v) = p(x)p(v\cond x)$, where $p(v\cond x)$ is an auxiliary distribution that provides another degree of freedom in the design of the kernel.
The key ingredients for choosing $p(v\cond x)$ are easy computation of its density and the ability to efficiently sample from it.

For the new target $p(x,v)$, we can apply the transition kernel \eqref{eq:det_kernel} as well as formulate Proposition \ref{th:x_cond}.
This can be done by simply rewriting these equations by substituting the tuple $[x,v]$ for the variable $x$.
Again, for the deterministic function $f(x,v)$, we resort to the family of involutive maps: $f(x,v) = f^{-1}(x,v)$.
However, in contrast to the case without the auxiliary variables, now we have an opportunity to reach any point of the target support by resampling $v\cond x$ before applying the deterministic map $f(x,v)$.
Interleaving the kernel \eqref{eq:det_kernel} with the resampling of $v$ one can collect samples from $p(x,v)$, and then obtain samples from the marginal distribution of interest $p(x)$ by simply ignoring $v$-coordinates of the collected samples.
We provide the pseudo-code in Algorithm \ref{alg:imcmc} below.
To get an intuition, one can think of the resulting algorithm as of a slightly abstract version of Hybrid Monte Carlo (HMC) \citep{duane1987hybrid}, where the momentum plays the role of the auxiliary variable $v$, and the Hamiltonian dynamics is a special case of the deterministic map $f$.

By construction, Algorithm \ref{alg:imcmc} keeps the joint density $p(x,v)$ invariant (satisfies the fixed point equation), but does not provide any guarantees for ergodicity.
In practice, the ergodicity is usually achieved by choosing proper involution and auxiliary distribution, such that the whole kernel is irreducible \citep{roberts2004general}.

\begin{algorithm}[H]
  \caption{Involutive MCMC}
  \begin{algorithmic}  
    \INPUT{target density $p(x)$}
    \INPUT{density $p(v\cond x)$ and a sampler from $p(v\cond x)$} 
    \INPUT{involutive $f(x,v)$, i.e. $f(x,v) = f^{-1}(x,v)$}
    \STATE initialize $x$
    \FOR{$i = 0\ldots n$}
        \STATE sample $v \sim p(v \cond x)$
        \STATE propose $(x',v') = f(x,v)$
        \STATE $P = \min\{1,\frac{p(x',v')}{p(x,v)}| \frac{\partial f(x,v)}{\partial [x,v]}|\}$
        \STATE $ x_{i} =
            \begin{cases} 
            x' , \text{ with probability } P\\
            x , \;\text{ with probability } (1-P)
            \end{cases}$
        \STATE $x \gets x_{i}$
    \ENDFOR
    \OUTPUT{samples $\{x_0,\ldots, x_n\}$}
  \end{algorithmic}
  \label{alg:imcmc}
\end{algorithm}
Among the kernels that satisfy the fixed point equation there is a family of kernels called \textit{reversible} which satisfy the detailed balance condition
\begin{align}
    t(x'\cond x)p(x) = t(x\cond x')p(x').
\end{align}
Such kernels are known to mix slower compared to the kernels that satisfy the fixed point equation but are \textit{irreversible} (do not satisfy the detailed balance condition) \citep{ichiki2013violation}.
In the following proposition, we demonstrate that the chain from Algorithm \ref{alg:imcmc} is reversible on both the support of $p(x,v)$ and $p(x)$ (proof in Appendix \ref{app:detailed_balance}).

\begin{proposition}
\label{th:detailed_balance}
Transition kernel $t(x',v'\cond x,v)$ from Algorithm \ref{alg:imcmc} satisfies detailed balance
\begin{align}
    t(x',v'\cond x,v) p(x,v) = t(x,v\cond x',v')p(x',v').
\end{align}
Moreover, the marginalized kernel on $x$
\begin{align}
    \widehat{t}(x'\cond x) = \int dvdv' t(x',v'\cond x,v)p(v\cond x)
    \label{eq:RT}
\end{align}
also satisfies detailed balance
\begin{align}
    \widehat{t}(x'\cond x) p(x) = \widehat{t}(x\cond x')p(x').
\end{align}
\end{proposition}

The reversibility of the chain $t(x',v'\cond x,v)$ is a direct consequence of the involutive property of the map $f(x,v)$ so it seems hard to avoid.
However, it is still possible to construct an irreversible chain by composing several reversible kernels.
We discuss this further in Section \ref{sec:compositions}.

\section{Tricks}
\label{sec:tricks}

The only two degrees of freedom possible to design in Involutive MCMC are the auxiliary distribution $p(v\cond x)$ and the involution $f(x,v)$.
However, we will show that many existent MCMC algorithms from the literature can be formulated as Involutive MCMC by choosing suitable $f(x,v)$ and $p(v\cond x)$. As such, iMCMC represents a unifying framework for understanding existing and designing new MCMC algorithms. 

We start by considering a simple involution $f(x,v) = [v,x]$ that is a swap of $x$ and $v$.
Choosing $q(v\cond x)$ such that the chain can reach any point in the support of $p(x)$, we end up with the Metropolis-Hastings algorithm (MH) with proposal $q(v\cond x)$.
Indeed, the acceptance probability in the Algorithm \ref{alg:imcmc} then equals
\begin{align*}
    P = \min\bigg\{1, \frac{p(f(x,v))}{p(x)q(v\cond x)}\bigg\}= \min\bigg\{1, \frac{p(v)q(x\cond v)}{p(x)q(v\cond x)}\bigg\}.
\end{align*}
While for MH the involution is very simple (a swap), we can also design MCMC algorithms by proposing sophisticated involutions. In the following subsections we explore this spectrum by demonstrating that a variety of MCMC algorithms can be formulated as Involutive MCMC methods.
To avoid the large amounts of technical details of all the considered algorithms, we formulate the most important ideas as \textit{tricks}.
Besides being the main ideas of the algorithms, these tricks can serve as useful tools to design efficient novel samplers.

\subsection{Smart auxiliary spaces}

In this subsection we consider algorithms that focus on the development of advanced auxiliary distributions.

We start with the trick that allows one to circumvent the evaluation of intractable integrals in the target distribution or in the auxiliary distribution when we use Algorithm \ref{alg:imcmc}.

\begin{trick}[Mixture distributions]
\label{th:mod}
Consider the joint distribution $p(x,v) = p(x)p(v\cond x)$, whose density is given as a mixture:
\begin{align}
    p(x) = \int p(x\cond z)p(z) dz, \\
    p(v\cond x) = \int q(v\cond a) q(a\cond x) da,
\end{align}
the evaluation of the integrals can be costly or even intractable.
One can bypass the integration by sampling from the joint distribution $p(x,v,z,a) = p(x\cond z) p(z) q(v\cond a) q(a\cond x)$ using the Algorithm \ref{alg:imcmc} with some involution $f(x,v,z,a)$.
Note that to sample $v$ we usually sample $a$ at each step; hence, we may leave this intermediate variable $a$ untouched by the subsequent involution, i.e. $f(x,v,z,a)=[f'(x,v,z),a]$. 
All arguments hold for discrete $a$ and $z$ as well.
Moreover, conditioning the distribution $q(v\cond a)$ by the current state $x$: $q(v\cond a,x)$, allows one to obtain a so-called state-dependent mixture as a proposal:
\begin{align}
    p(v\cond x) = \int q(v\cond a,x) q(a\cond x) da.
\end{align}
In the discrete case
\begin{align}
    p(v\cond x) = \sum_j q(v\cond j,x) q(j\cond x).
\end{align}
\end{trick}

This trick immediately implies the algorithm proposed in \citep{habib2018auxiliary} (see Appendix \ref{app:MPMCMC} for the proof), where the authors consider the proposal $q(x'\cond x)$ for the Metropolis-Hastings algorithm as a mixture $q(x'\cond x) = \int q(x'\cond a) q(a\cond x)da$.
Another application of this trick can be found in the Multiple-Try Metropolis scheme \citep{liu2000multiple} and Reversible-Jump MCMC \citep{green1995reversible}.
We will return to these algorithms shortly.

Note, however, that avoiding by this trick the analytical integration, one only shifts the integration burden to the algorithm reducing its efficiency.
Indeed, extending the target distribution with additional variables requires the sampling in higher dimensions, which may result in a slower convergence and a higher variance of the estimate.

The mixture of auxiliary variables in Trick \ref{th:mod} can be considered as an adaptive change of the family of proposal distributions depending on the current state of the chain.
Another way to enrich the set of proposed points is to choose a suitable involution based on the current state.
We describe this idea in the following trick.

\begin{trick}[Mixture of Involutions]
\label{th:moi}
Consider the joint distribution $p(x,v) = p(x)p(v\cond x)$ and a parametric family of involutions $f_a(x,v) = f^{-1}_a(x,v)$, i.e., functions that define a proper involution in the space of tuples $[x,v]$ for a given $a$.
It can be useful to apply different involutive maps depending on the current state $[x,v]$.
For that purpose, one may introduce an auxiliary random variable $a$ and define the joint distribution $p(x,v,a) = p(x,v)p(a\cond x,v)$.
Then the involution in the new space is $f'(x,v,a) = [f_a(x,v),a]$, and the acceptance probability is
\begin{align*}
    P = \min\bigg\{1,\frac{p(f_a(x,v))p(a\cond f_a(x,v))}{p(x,v)p(a\cond x,v)} \bigg|\frac{\partial f_a(x,v)}{\partial [x,v]}\bigg|\bigg\}.
\end{align*}
\end{trick}
We thus observe that by first sampling $v\sim p(v\cond x)$ and $a\sim p(a\cond x,v), $ and then applying $f_a(x,v)$ we can have different involutions depending on $x,v$. 

The crucial part of this trick is leaving the auxiliary variable $a$ invariant by the involution in order to satisfy the fixed point equation. The correctness of such a kernel can be obtained by application of Proposition \ref{th:x_cond} for the tuple $[x,v,a]$ and the target distribution $p(x,v,a)$.
Moreover, we immediately obtain the reversibility of this kernel from Proposition \ref{th:detailed_balance}.
For more details and formal derivations, we refer the reader to Appendix \ref{app:moi_proof}.

Together with Trick \ref{th:mod}, this trick provides an iMCMC formulation of the Multiple-Try Metropolis scheme \citep{liu2000multiple}.
Speaking informally, we generate several proposals $v$, indexed by the variable $a$, using the mixture of distributions from Trick \ref{th:mod}, and then stochastically decide which swap we use to propose the next state (see Appendix \ref{app:MTM} for the proof).

Surprisingly, using this trick we obtain the iMCMC formulation of Sample-Adaptive MCMC \citep{zhu2019sample} (see Appendix \ref{app:SAMCMC}), which does not have the MH acceptance test at all.
Furthermore, Sample-Adaptive MCMC greatly relies on the aggregation functions that do not depend on the order of their arguments, i.e.
\begin{align}
    g(x_1,\ldots, x_n) = g(\pi(x_1,\ldots, x_n)),
\end{align}
where $\pi(\cdot)$ is an arbitrary permutation.
Using the iMCMC formalism we can easily remove this restriction and obtain a more general scheme (see Appendix \ref{app:gSAMCMC}).

Further, we will use Trick \ref{th:moi} for the reformulation of several algorithms: Reversible-Jump MCMC \citep{green1995reversible}, Non-Reversible Jump scheme \citep{gagnon2019non}, and Look Ahead HMC \citep{sohl2014hamiltonian}.

\subsection{Smart deterministic maps}

In this subsection we consider algorithms that introduce sophisticated involutive maps to obtain an efficient sampler.
Historically, the first example is the Hybrid Monte Carlo (HMC) algorithm \citep{duane1987hybrid}.
Its core part is the Leap-Frog integrator of the corresponding Hamiltonian dynamics.
We denote a single application of Leap-Frog as $L$ and its iterative application as $L^k$, where $k$ is the number of applications (steps of the dynamics).
Then we can formulate HMC in terms of Involutive MCMC as follows.
Consider the joint distribution $p(x,v) = p(x)p(v)$, where $p(v)$ usually equals to the standard normal and $v$ represents the momentum variable.
The involutive map can be constructed as the composition $FL^k$, where $F$ denotes the momentum flip operator ($F(x,v) = [x,-v]$), and is applied after $k$ iterative applications of $L$.
According to the iMCMC formalism, the acceptance probability in Algorithm \ref{alg:imcmc} is
\begin{align}
    P = \min \bigg\{1, \frac{p(FL^k(x,v))}{p(x,v)} \bigg\}.
\end{align}
Here we use the fact that both $L$ and $F$ preserves volume, hence, their Jacobians equal to $1$.
For the formal proof see Appendix \ref{app:HMC}.

Contrary to the MH algorithm, for HMC we see that all the "knowledge about the target" of a sampler is concentrated in the involutive map. In contrast, the distribution $q(v|x)$ is very simple (a standard normal independent of $x$).
This fact motivates the number of MCMC algorithms that try to build expressive deterministic maps using neural networks. We describe the main ideas of these algorithms in the following tricks.

\begin{trick}[Auxiliary direction]
\label{th:direction}

Consider the joint distribution $p(x,v) = p(x)p(v\cond x)$, which we denote as $p(y)=p(x,v)$ with $y=[x,v]$.
To obtain an expressive sampler, one can construct the required involution $f$ using some non-involutive bijection $T(y)$ in the following way.

Consider the joint distribution $p(y,d) = p(y)p(d\cond y)$, where the binary auxiliary variable $d=\{-1,+1\}$ encodes the direction in which we move from the current state.
The involution $f$ is then constructed as $f(y,d=+1) = [T(y), -1], \; f(y,d=-1) = [T^{-1}(y), +1]$.
The acceptance probability is
\begin{align}
    P = \min\bigg\{1,\frac{p(T_d(y))p(-d\cond T_d(y))}{p(y)p(d\cond y)} \bigg|\frac{\partial T_d}{\partial y}\bigg| \bigg\},
\end{align}
where $T_{d=+1} = T$, and $T_{d=-1} = T^{-1}$.

More generally, we can choose $d$ to lie in a vector space, parameterizing the family of bijections $T_d(y)$.
To construct an involution we require that for any $d$ there exists a unique $d'$ such that $T_{d'} = T_d^{-1}$ and a smooth map $g(d) = d'$.
Note that by requiring $T_{d'} = T_d^{-1}$ we immediately obtain $T_{d'}^{-1} = T_d$, hence $g(d') = d$ meaning that $g$ is an involution.
The final involution is then $f(y,d) = [T_d(y), g(d)]$, and the acceptance probability is
\begin{align}
    P = \min\bigg\{1,\frac{p(T_d(y), g(d))}{p(y,d)} \bigg|\frac{\partial T_d}{\partial y}\bigg| \bigg|\frac{\partial g}{\partial d}\bigg| \bigg\},
\end{align}
For instance, one can consider a Lie group on $\mathbb{R}^n$ with the group operation $T(y,d) = yd: \mathbb{R}^n\times\mathbb{R}^n \to \mathbb{R}^n$.
Then the involution $f$ can be constructed as $f(y,d) = [yd, d^{-1}]$.
\end{trick}

The described trick is a generalization of A-NICE-MC algorithm \citep{song2017nice} (see Appendix \ref{app:nicemc}), and L2HMC algorithm \citep{levy2017generalizing} (see Appendix \ref{app:l2hmc}).
Indeed, considering the uniform distribution over the binary auxiliary variable $p(d\cond x,v) = p(d) = \text{Uniform}\{-1,+1\}$, and taking the bijection $T(x,v)$ as the corresponding model, we immediately obtain both algorithms.
Furthermore, taking the same distribution for the auxiliary direction $p(d) = \text{Uniform}\{-1,+1\}$ and combining Tricks \ref{th:mod}, \ref{th:moi}, \ref{th:direction} we formulate Reversible-Jump MCMC algorithm \citep{green1995reversible} in terms of iMCMC (see Appendix \ref{app:RJMCMC}).

Note that also vanilla HMC can easily be expressed using this trick by using $T(y) = L(y)$ and $T^{-1}(y) = L^{-1}(y) = FLF(y)$ and where $d \sim \text{Uniform}\{-1,+1\}$.
Indeed, the flips $F(y = [x,v]) = [x,-v]$ in the Leap-Frog procedure do not influence the chain when $p(x,v) = p(x,-v)$.
We will return to this formulation of HMC during the discussion of irreversible chains.

In the following trick, we demonstrate another way to use a bijection $T$ to construct an expressive involution that generalizes the NeuTra algorithm of \citep{hoffman2019neutra} (see Appendix \ref{app:neutra}).

\begin{trick}[Embedded involution]
\label{th:embed_involution}

Consider the iMCMC sampler with the joint distribution $p(x,v) = p(x)p(v\cond x)$ and the involution $f(x,v) = f^{-1}(x,v)$.
Assume that for some reason the involution $f$ is not expressive enough to yield an efficient sampler for the target distribution $p(x)$.
We can enrich the sampler by choosing a suitable bijection $T$, and introducing the new involution $f_T = T^{-1}\circ f \circ T$, where $\circ$ is the composition operation.

Moreover, consider the embedded random variable $[X_T,V_T] = T(X,V), \; [X,V] \sim p(x,v)$ with the density 
\begin{align}
    p_T(x_T,v_T) = p(T^{-1}(x_T,v_T)) \bigg|\frac{\partial T^{-1}}{\partial [x_T,v_T]}\bigg|.
\end{align}
Then the Algorithm \ref{alg:imcmc} with the joint distribution $p(x,v)$ and the involution $f_T$ is equivalent to the following procedure.
Given the sample $[x,v] \sim p(x,v)$, map this sample as $T(x,v)$.
Starting from $T(x,v)$, collect new samples $\{(x_T,v_T)_i\}$ from $p_T(x_T,v_T)$ using the Algorithm \ref{alg:imcmc} with the joint distribution $p_T(x_T,v_T)$ and the involution $f(x_T,v_T)$. 
Then map all the collected samples as $T^{-1}(x_T,v_T)$.
\end{trick}

The map $T$ can be viewed as a 'flow' model to a simpler 'disentangled' or more symmetric latent space $x_T,v_T$  where algorithms such as HMC are easier to run. The map $T$ could be learned using unsupervised learning on already generated samples.  

\subsection{Smart compositions}
\label{sec:compositions}

In this subsection, we apply the formalism of involutive MCMC to describe irreversible chains, i.e. chains that do not satisfy detailed balance. 
Recall that in Section \ref{sec:iMCMC} we have shown that any iMCMC chain must be reversible.
However, a composition of reversible chains is not necessarily reversible. 
Thus, in the following tricks, we use reversible iMCMC chains as building blocks to construct a composition that is irreversible.
A representative example of such a composition is Gibbs sampling \citep{geman1984stochastic}.
Indeed, the update of a single coordinate in the Gibbs algorithm is a reversible kernel easily described by the iMCMC framework, while the composition of these kernels yields Gibbs sampling which is irreversible (see Appendix \ref{app:gibbs}).


Using a composition of kernels, we can make an irreversible analogue of Trick \ref{th:direction}. The main difference is that in this Trick we do not resample the auxiliary (directional) variable $d$ at every iteration. This would reverse the direction after each accepted proposal. However, by composing this with a kernel that simply flips $d$ again we get a persistent (irreversible) kernel that only flips directions when a sample is rejected. 

\begin{trick}[Persistent direction]
\label{th:persistent_dir}

Given the target distribution $p(y) = p(x,v)$, we consider the joint distribution $p(y,d) = p(y)p(d)$, where $p(d)=\text{Uniform}\{-1,+1\}$, and the variable $d=\{-1,+1\}$ encodes the direction in which we move from the current state.
Following Trick \ref{th:direction}, we consider some non-involutive bijection $T(y)$ and the corresponding involution $f_1(y,d=+1) = [T(y), -1], \; f_1(y,d=-1) = [T^{-1}(y),+1]$.
Thus, we obtain the iMCMC kernel $t_1(y',d'\cond y,d)$ that accepts the proposal point $[T_d(y),-d]$ with the probability
\begin{align}
    P_1 =& \min\bigg\{1,\frac{p(T_d(y))p(-d)}{p(y)p(d)} \bigg|\frac{\partial T_d}{\partial y}\bigg| \bigg\}= \\
    =&\min\bigg\{1,\frac{p(T_d(y))}{p(y)} \bigg|\frac{\partial T_d}{\partial y}\bigg| \bigg\},
\end{align}
where $T_{d=+1} = T$, and $T_{d=-1} = T^{-1}$.
Then we compose the kernel $t_1$ with the kernel $t_2$ that just flips the directional variable.
In terms of iMCMC, the target distribution is $p(y,d)$ and the involution is $f_2(y,d) = [y,-d]$.
Note that this proposal will be always accepted since $p(y,-d) = p(y,d)$.
Then the composition of $t_1$ and $t_2$ works as follows.
\begin{align}
    &\text{current state} = [y,d]\\
    &\text{next state} =
    \begin{cases}
    [T_d(y), d], \text{ with probability } P_1 \\
    [y,-d], \text{ with probability } (1-P_1)
    \end{cases}
\end{align}
The same logic can be applied to the variable $v$.
Since $T_d(y) = T_d(x,v)$ may significantly depend on the variable $v$, instead of resampling it at each step, one can use another kernel to update it conditioned on its previous value.
\end{trick}

\begin{figure}[h]
    \centering
    \includegraphics[width=0.4\textwidth]{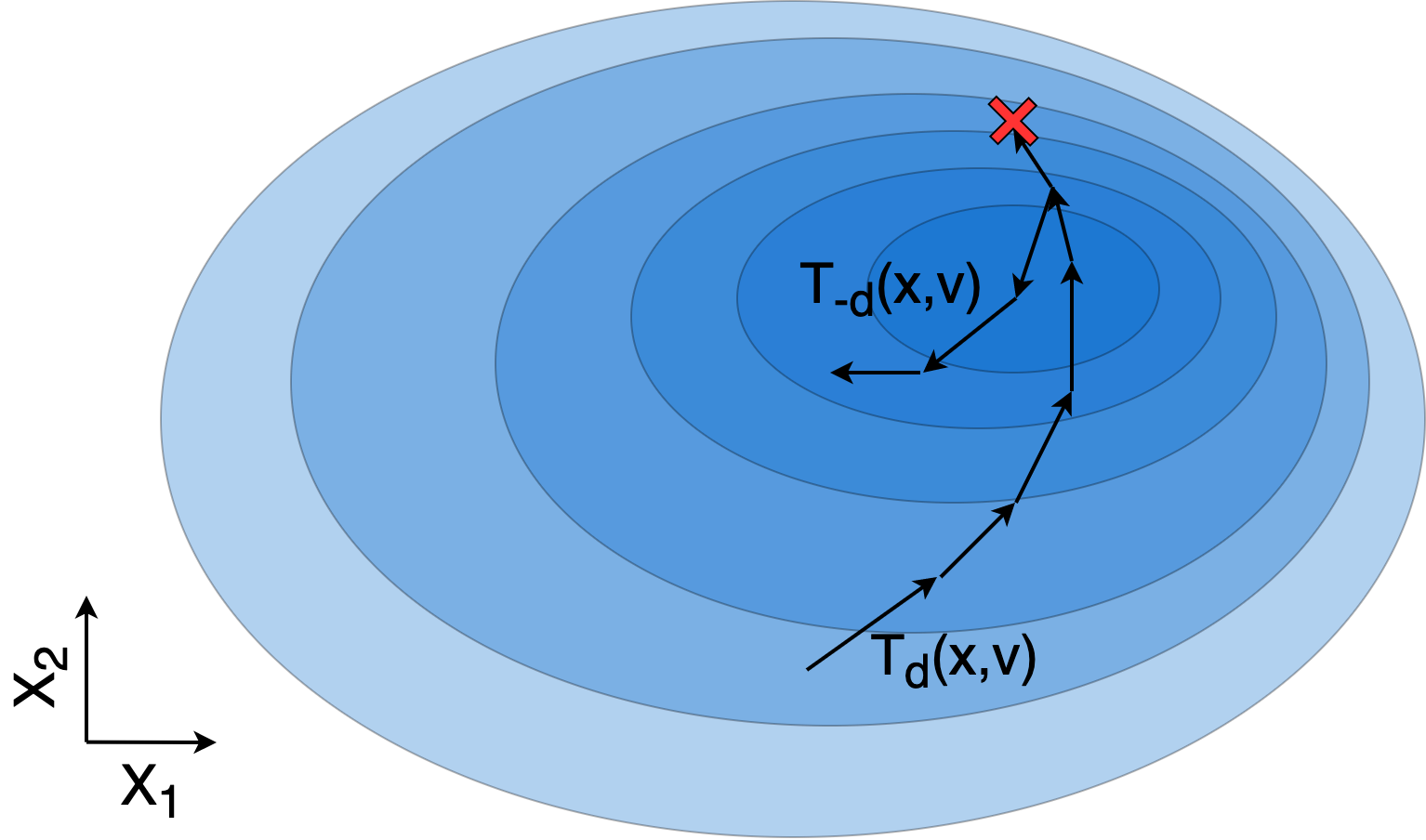}
    \caption{Schematic representation of Trick \ref{th:persistent_dir}.
    The deterministic map $T_d(y) = T_d(x,v)$ is iteratively applied with resampling of the variable $v$, moving in the regions of a high density.
    When the chain tries to move to a lower density region, the proposal may be rejected (red cross), which triggers the chain to change direction and apply $T_{-d}(x,v)$.}
    \label{fig:persistent_dir}
\end{figure}

The intuition of this composition is as follows. 
In the case of an accept we now try to move further by applying the same $T_d$ instead of the inverse map $T_{-d}$, whereas, in the case of a reject, we flip the variable $d$ and move in the opposite direction.
We depict this intuition in Fig. \ref{fig:persistent_dir}.

Trick \ref{th:persistent_dir} describes the main idea behind persistent chains leading to irreversibility.
Its generalizations can be derived similar to what we explained in Trick \ref{th:direction}, by considering a conditional direction $p(d\cond y)$ or a vector valued direction variable $d$. 
If we use a distribution $p(d\cond y)$, we must also take care to change the second kernel $t_2$ to preserve the target $p(y,d)$.

The analogue of the direction flip in Trick \ref{th:persistent_dir} may be found in the HMC algorithm with persistent momentum \citep{horowitz1991generalized} (see Appendix \ref{app:horowitz}) and the Look Ahead HMC algorithm \citep{sohl2014hamiltonian} (see Appendix \ref{app:LAHMC}). 
These algorithms use post-acceptance negation of the momentum variable $v$ relying on the symmetry of the auxiliary distribution: $p(-v) = p(v)$.
However, using Trick \ref{th:persistent_dir} we can easily generalize these algorithms to the case of an asymmetric auxiliary distribution $p(-v) \neq p(v)$ by considering the forward map as a Leap-Frog operator $T_{d=+1}(y)=L(y)$ and its inverse as $T_{d=-1}(y)=L^{-1}(y)$, where $L^{-1}$ is the Leap-Frog backward in time.
More details are provided in Appendix \ref{app:horowitz}.

In light of Tricks \ref{th:moi} and \ref{th:persistent_dir}, we can obtain yet another generalization of Look Ahead HMC \citep{sohl2014hamiltonian}.
In this paper, the authors propose the mixture of involutions $f_k(x,v) = FL^k(x,v)$, where we choose $k$ stochastically based on the current state.
Using Trick \ref{th:persistent_dir}, we can look ahead of any function we want, by considering the family of involutions 
\begin{align}
    f_k(x,v,d=+1) = [T^k(x,v),-1], \\
    f_k(x,v,d=-1) = [T^{-k}(x,v),+1],
\end{align}
where $T^k$ means $k$ iterative applications of the map $T$, and $T^{-k}$ means the same but for the map $T^{-1}$.
Note that by considering $T = L$ and $T^{-1} = FLF$, and the symmetric auxiliary distribution $p(v)=p(-v)$ we obtain Look Ahead HMC.

The combination of Tricks \ref{th:moi}, \ref{th:persistent_dir} provides a neat iMCMC formulation of Gibbs sampling \citep{geman1984stochastic} and Non-Reversible Jump MCMC \citep{gagnon2019non}. Details can be found in appendices \ref{app:gibbs} and \ref{app:nrj} respectively.

\begin{figure}[h]
    \centering
    \includegraphics[width=0.3\textwidth]{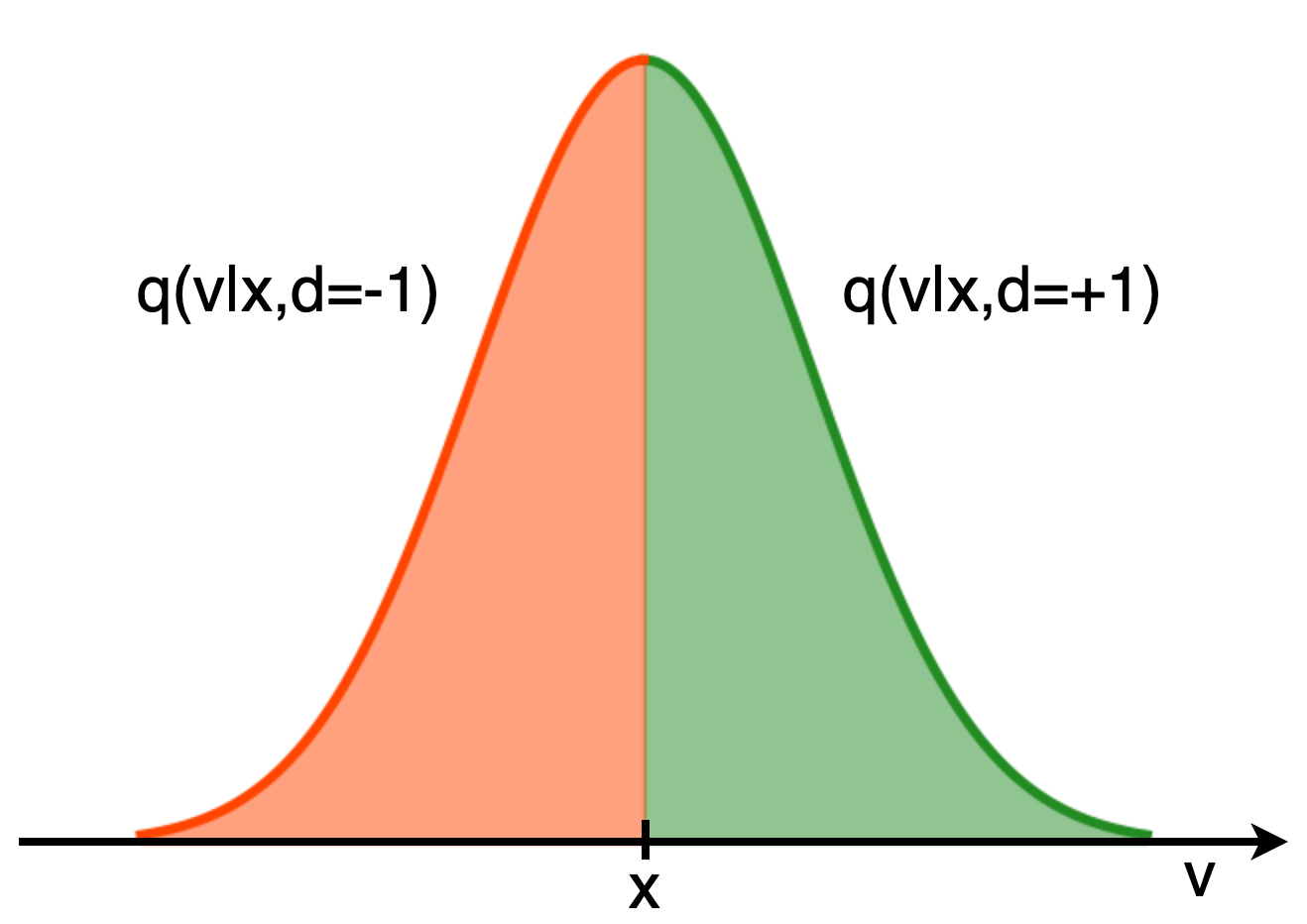}
    \caption{Schematic representation of auxiliary distribution in Trick \ref{th:lifting}.
    Red mass represents $q(v\cond x,d=-1)$, green mass represents $q(v\cond x,d=+1)$.
    These proposals do not intersect only for illustrative purposes to highlight that we cannot move to the left if $d=+1$.}
    \label{fig:lifting}
\end{figure}

Trick \ref{th:persistent_dir} tells us how to design an irreversible chain using the deterministic part of the iMCMC framework.
The following trick makes it possible using the stochastic auxiliary variables.

\begin{trick}[Persistent proposal]
\label{th:lifting}
Consider the joint target distribution $p(x,d)$, where $p(d)=\text{Uniform}\{-1,+1\}$ is the directional variable.
Choose the auxiliary distribution $q(v\cond x, d)$ that proposes new points depending on the current direction $d$.
For instance, this can be done by splitting a random walk proposal $q(v\cond x)$ as depicted in Fig. \ref{fig:lifting}.
Then the constructed iMCMC kernel $t_1(x',d'\cond x,d)$ with involution $f(x,v,d) = [v,x,-d]$ has the probability of acceptance:
\begin{align}
    P_1 = \min \bigg\{1,\frac{p(v)q(x\cond v,-d)}{p(x) q(v\cond x,d)} \bigg\}.
\end{align}
Note that having proposals as depicted in Fig. \ref{fig:lifting} we must change the directional variable when proposing the next state, otherwise we obtain zero probability of acceptance.

\begin{table*}[t]
\caption{
Performance of the algorithms as measured by the batch-means estimator of Effective Sample Size (ESS) averaged across $100$ independent chains. 
Higher values of ESS and ESS per second are better (for detailed formulation see Appendix \ref{app:ess}).
For computational efforts we provide ESS per second taking into account the sampling time for $20000$ samples.
See description of the compared methods in the text.}
\label{tab:examples}
\vspace{-1em}
\begin{center}
\resizebox{1.0\textwidth}{!}{
\begin{tabular}{lcccc|cccc}
    \toprule
    \multicolumn{1}{c}{} & \multicolumn{4}{c}{\bf ESS} & \multicolumn{4}{c}{\bf ESS per second}\\
    \midrule
    {Algorithm}&{MoG2}&{Australian}&{German}&{Heart}&{Mog2}&{Australian}&{German}&{Heart}\\
    \midrule
    MALA & $0.007\pm0.002$ & $\mathbf{0.043\pm0.001}$ & $\mathbf{0.025\pm0.005}$ & $\mathbf{0.081\pm0.012}$ & $2$ & $5$ & $3$ & $9$ \\
    Irr-MALA & $\mathbf{0.027\pm0.008}$ & $0.006\pm0.001$ & $0.004\pm0.001$ & $0.012\pm0.001$ & $4$ & $1$ & $1$ & $1$ \\
    \hline
    A-NICE-MC & $0.852\pm0.239$ & $0.137\pm0.026$ & $0.032\pm0.004$ & $0.253\pm0.033$ & $1700$ & $94$ & $17$ & $241$ \\
    Irr-NICE-MC & $\mathbf{1.643\pm0.626}$ & $\mathbf{0.177\pm0.030}$ & $0.032\pm0.004$ & $\mathbf{0.341\pm0.051}$ & $\mathbf{3280}$ & $\mathbf{121}$ & $17$ & $\mathbf{324}$ \\
    \bottomrule
\end{tabular}}
\end{center}
\end{table*}

As well as in Trick \ref{th:persistent_dir}, we then compose the kernel $t_1$ with the kernel $t_2$ that just flips the directional variable.
In terms of iMCMC that is, the target distribution is $p(x,d)$ and the involution is $f_2(x,d) = [x,-d]$.
Note that this proposal will be always accepted since $p(x,-d) = p(x,d)$.
Then the composition of $t_1$ and $t_2$ works as follows.
\begin{align}
    &\text{current state} = [x,d]\\
    &\text{proposal} = v \sim q(v\cond x,d)\\
    &\text{next state} =
    \begin{cases}
    [v, d], \text{ with probability } P_1 \\
    [x,-d], \text{ with probability } (1-P_1)
    \end{cases}
\end{align}

Once again, a more general version of this trick can be obtained as in Trick \ref{th:direction}, by considering a conditional direction $p(d\cond x)$ or a direction vector-valued $d$. 
If we change the distribution to $p(d\cond x)$, we must also change the second kernel $t_2$ to preserve the target $p(x,d)$.
\end{trick}

Implicitly this trick is used in the Lifted Metropolis-Hastings algorithm \citep{turitsyn2011irreversible}, which gives a rise to many irreversible algorithms.
The only difference with Trick \ref{th:lifting} is that Lifted MH design the proposal distribution $q(v\cond x,d)$ as the transition kernel of the conventional MH algorithm (see Appendix \ref{app:lmh}).

Note that taking the kernels $t_{+}$ and $t_{-}$ that already satisfy the generalized detailed balance $t_{+}(x'\cond x)p(x) = t_{-}(x\cond x')p(x')$ as positive and negative parts of $q(x\cond v, d)$, we obtain the irreversible chain that is equivalent to the application either of $t_{+}$ or $t_{-}$.

\section{Examples}
\label{sec:applications}

We now proceed with illustrating that the proposed framework provides an easy paradigm to extend and combine existing methods and potentially improve them.
Below, we propose simple extensions, to make MALA and A-NICE-MC irreversible. We empirically validate these examples on a set of tasks, which includes a mixture of two 2d-Gaussians (MoG2) and the posterior distribution of Bayesian logistic regression on several datasets (Australian, German, Heart) (see Appendix \ref{app:dists} for details).
For performance evaluation, we use the effective sample size (ESS), which measures how many independent samples the chain actually contains.
To be more precise, we use batch-means estimator of ESS, which is shown to be more robust \citep{thompson2010comparison} (see Appendix \ref{app:ess} for details).

We start with the Metropolis-Adjusted Langevin algorithm (MALA) \citep{roberts1996exponential}, which generates proposals by following the gradient of the target log-probability.
We modify the original algorithm with a directional variable $p(d) = \text{Uniform}\{-1,+1\}$ as follows.
The joint distribution is now:
\begin{align*}
    p(x,v,d) = p(x)\Normal(v\cond x + d\eps \nabla_x\log p(x), 2\eps)p(d),
\end{align*} 
and the involutive map is 
\begin{align*}
    f(x,v,d) = [v,x,-d\cdot\text{sign}(\nabla_x\log p(x)^T\nabla_v\log p(v))].
\end{align*}
Thus, our modification (Irr-MALA) ensures that the gradient in the proposed point will be directed towards the initial point.
The irreversible chain can be obtained by the application of the described kernel followed by the negation of $d$ (see Appendix \ref{app:neg-mala} for pseudo-code).
However, allowing the chain to traverse along the gradient of decreasing probability reduces its acceptance rate, which leads to a poor performance on the unimodal posteriors of Bayesian logistic regression.
However, if we need to traverse low probability regions between two modes of a distribution this idea becomes beneficial and leads to improved performance. We can see this when we sample from the bimodal MoG2 distribution. (see Table \ref{tab:examples}).

We now turn to a more complex model, and design the irreversible version of A-NICE-MC \citep{song2017nice}, which learns the NICE model \citep{dinh2014nice} to obtain an expressive proposal.
Our modification (Irr-NICE-MC) is a straightforward application of Trick \ref{th:persistent_dir} to the original algorithm (see Appendix \ref{app:irr-nice} for pseudo-code).
The only difference with Trick \ref{th:persistent_dir} is that we add one more kernel that conditionally updates the auxiliary variable $v$ as
\begin{align}
    v' = v\sqrt{1-\alpha^2} + \alpha\cdot\eta, \;\; \eta \sim \Normal(0,1).
\end{align}
For all targets we choose $\alpha=0.8$.
To provide a robust comparison, we do not change the training process of the original algorithm. 
Moreover, we compare our modification against the original method, using the same--already learned--model as the proposal distribution.
In Table \ref{tab:examples}, we see that simply introducing irreversibility into the kernel may result in significant performance gains while having a negligible computational overhead. 

Code for reproducing the experiments is available at \url{https://github.com/necludov/iMCMC}.

\section{Related work}

Several approaches unifying MCMC exist in the literature.
They focus on the kernels with multiple proposals and describe them, extending the state space through the auxiliary variables \citep{tjelmeland2004using,storvik2011flexibility}.
The most general unifying framework is given in \citep{finke2015extended}, which considers all Monte Carlo algorithms as the importance sampling on differently extended spaces.

The key difference of the proposed framework is the explicit usage of the involutive deterministic map inside of the generic kernel.
Although this deterministic part appears to be trivial in some cases (for instance, the swap in the MH algorithm), it may serve as a design principle for many MCMC algorithms.
The main benefit of this principle comes when we consider hybrid algorithms that incorporate expressive deterministic maps into MCMC kernels.
Such hybrid algorithms demonstrate promising results in modern physics \citep{kanwar2020equivariant}, and the iMCMC framework may give a hint on how to design these algorithms as well as how to combine features of different MCMC kernels.

\section{Conclusion}
In this paper, we have proposed a unifying view of a large class of MCMC algorithms. This was achieved by reformulating MCMC as the composition of sampling from an auxiliary distribution and an involutive deterministic map, followed by a MH accept-reject step. This was shown to represent a very large family of reversible MCMC algorithms. We then extend this class further with the use of auxiliary variables into irreversible MCMC algorithms. Through a number of ``Tricks'' we facilitate the process of extending existing algorithms, which we illustrate through some simple examples. 

We believe our unifying view of MCMC algorithms will lead to a number of generalizations in the future. For instance, some of the versions look very similar to flow-based models used for unsupervised learning and some of our proposed kernels can indeed be used for such a purpose. We also believe there are interesting connections between deterministic samplers as in \citep{murray2012driving, neal2012view} and the theory of (chaotic) iterated maps and nonlinear dynamical systems. 

\section{Acknowledgments}

The authors are thankful to the reviewers who provided detailed feedback and pointed out essential related works.
Kirill Neklyudov and Dmitry Vetrov have been supported by the Russian Science Foundation grant no.~19-71-30020.

\bibliography{icml2020}
\bibliographystyle{icml2020}

\newpage
\onecolumn

\appendix

\section{Involutive MCMC}

\subsection{Proof of Proposition \ref{th:x_cond} (FPE condition)}
\label{app:fpe_condition}

For the target distribution $p(x)$ and the deterministic proposal $q(x'\cond x) = \delta (x' - f(x))$, we consider the following transition kernel
\begin{align}
    t(x'\cond x) =  \; \delta (x' - f(x)) \min \bigg\{1, \frac{p(x')}{p(x)}\bigg| \frac{\partial f}{\partial x} \bigg|\bigg\} + \delta(x'-x) \int dx''\; \delta (x'' - f(x))\bigg(1 - \min\bigg\{1, \frac{p(x'')}{p(x)} \bigg| \frac{\partial f(x)}{\partial x} \bigg| \bigg\}\bigg).
\end{align}
Then we want to check the fixed-point equation
\begin{align}
    \int dx\; t(x'\cond x)p(x) = p(x').
\end{align}
Substitution of $t(x'\cond x)$ gives
\begin{align}
    \int dx\;\delta (x' - f(x))\min\bigg\{p(x),p(x') \bigg| \frac{\partial f}{\partial x} \bigg| \bigg\} + p(x') - \min\bigg\{p(x'), p(f(x')) \bigg| \frac{\partial f}{\partial x'} \bigg| \bigg\} = p(x')
\end{align}
Assuming that $f(x)$ has the inverse $f^{-1}(x)$, we change variables $x = f^{-1}(\widetilde{x})$ and rewrite the previous equation as
\begin{align}
    \int d\widetilde{x}\; \delta (x' - \widetilde{x}) \min\bigg\{p(f^{-1}(\widetilde{x})),p(x') \bigg|\frac{\partial f}{\partial x}\bigg|_{x=f^{-1}(\widetilde{x})}\bigg\} \bigg|\frac{\partial f^{-1}}{\partial \widetilde{x}}\bigg| - \min\bigg\{p(x'), p(f(x')) \bigg| \frac{\partial f}{\partial x'} \bigg| \bigg\} = 0.
\end{align}
Using the chain rule, we have
\begin{align}
    1 = \bigg|\frac{\partial f(f^{-1}(x))}{\partial x}\bigg| = \bigg|\frac{\partial f}{\partial y}\bigg|_{y=f^{-1}(x)} \bigg|\frac{\partial f^{-1}}{\partial x}\bigg|.
\end{align}
Thus, we obtain the following condition to satisfy the fixed-point equation
\begin{align}
    \min\bigg\{p(f^{-1}(x))\bigg|\frac{\partial f^{-1}}{\partial x}\bigg| ,p(x)\bigg\} = \min\bigg\{p(x), p(f(x)) \bigg| \frac{\partial f}{\partial x} \bigg| \bigg\}.
\label{eq:cond}
\end{align}
The same applies for the joint space
\begin{align}
    \min\bigg\{p(f^{-1}(x,v))\bigg|\frac{\partial f^{-1}(x,v)}{\partial [x,v]}\bigg| ,p(x,v)\bigg\} = \min\bigg\{p(x,v), p(f(x,v)) \bigg| \frac{\partial f(x,v)}{\partial [x,v]} \bigg| \bigg\}.
\end{align}
Moreover, there is no need to care about the distribution of $v'$ in the fixed point equation
\begin{align}
    \int dxdvdv'\; t(x',v'\cond x,v)p(x,v) = p(x').
\end{align}
Thus, we obtain more general condition
\begin{align}
    \int dv \min\bigg\{p(f^{-1}(x,v))\bigg|\frac{\partial f^{-1}(x,v)}{\partial [x,v]}\bigg| ,p(x,v)\bigg\} = \int dv \min\bigg\{p(x,v), p(f(x,v)) \bigg| \frac{\partial f(x,v)}{\partial [x,v]} \bigg| \bigg\}.
\label{eq:gen_cond}
\end{align}
Also, note that the condition can be easily rewritten for different acceptance function, e.g., for the Barker's test \citep{barker1965monte}.
That is,
\begin{align}
    t(x'\cond x) =  \; \delta (x' - f(x)) \bigg[1+\frac{p(x)}{p(x')}\bigg| \frac{\partial f}{\partial x} \bigg|^{-1}\bigg]^{-1} + \delta(x'-x) \bigg(1 - \bigg[1+\frac{p(x)}{p(f(x))}\bigg| \frac{\partial f}{\partial x} \bigg|^{-1}\bigg]^{-1}\bigg).
\end{align}
Substituting this kernel into the fixed point equation $\int dx t(x'\cond x)p(x) = p(x')$, and performing a similar algebra, we have
\begin{align}
    \bigg[\frac{1}{p(x)}+\frac{1}{p(f^{-1}(x))}\bigg| \frac{\partial f^{-1}}{\partial x} \bigg|^{-1}\bigg]^{-1} = \bigg[\frac{1}{p(x)}+\frac{1}{p(f(x))}\bigg| \frac{\partial f}{\partial x} \bigg|^{-1}\bigg]^{-1}
\end{align}
Thus, for the Barker's test, the fixed point equation can be reduced to
\begin{align}
    p(f^{-1}(x))\bigg|\frac{\partial f^{-1}}{\partial x}\bigg| = p(f(x))\bigg|\frac{\partial f}{\partial x}\bigg|
\end{align}

\subsection{Proof of Proposition \ref{th:detailed_balance} (Detailed balance)}
\label{app:detailed_balance}

We analyse this property of Involutive MCMC by deriving the reverse operator $r(x,v\cond x',v')$, which is defined as
\begin{align}
    t(x',v'\cond x,v) p(x,v) = r(x,v\cond x',v')p(x',v').
\end{align}
By the definition, we have
\begin{align}
    r(x,v \cond x',v') =& t(x',v' \cond x,v) \frac{p(x,v)}{p(x',v')}\\
    r(x,v \cond x',v') =& \delta ([x',v'] - f(x,v)) \min \bigg\{\frac{p(x,v)}{p(x',v')}, \bigg| \frac{\partial f(x,v)}{\partial [x,v]} \bigg|\bigg\} + \\
    & + \delta([x',v']-[x,v]) \bigg(\frac{p(x,v)}{p(x',v')} - \min\bigg\{\frac{p(x,v)}{p(x',v')}, \frac{p(f(x,v))}{p(x',v')} \bigg| \frac{\partial f(x,v)}{\partial [x,v]} \bigg| \bigg\}\bigg)
\end{align}
The detailed balance is satisfied in the joint space if $\int_A r(x,v\cond x',v')dxdv = \int_A t(x,v\cond x',v') dxdv$, where $A$ is any non-zero measure volume in the joint space.
Remind that
\begin{align}
    t(x,v\cond x',v') =& \delta ([x,v] - f(x',v')) \min \bigg\{1, \frac{p(x,v)}{p(x',v')}\bigg| \frac{\partial f(x',v')}{\partial [x',v']} \bigg|\bigg\} + \\ 
    & + \delta([x,v]-[x',v']) \bigg(1 - \min\bigg\{1, \frac{p(f(x',v'))}{p(x',v')} \bigg| \frac{\partial f(x',v')}{\partial [x',v']} \bigg| \bigg\}\bigg).
\end{align}
For the involutive map $f$, it is clear that the integrals $\int_A r(x,v\cond x',v')dxdv$ and $\int_A t(x,v\cond x',v')dxdv$ are non-zero around the points $[x,v] = [x',v']$ and $[x,v] = f(x',v')$.
Thus, integrating over $A_1$ that is around $[x,v]=[x',v']$, we have
\begin{align}
    \int_{A_1} r(x,v\cond x',v')dxdv = 1 - \min\bigg\{1, \frac{p(f(x',v'))}{p(x',v')} \bigg| \frac{\partial f(x',v')}{\partial [x',v']} \bigg| \bigg\} = \int_{A_1} t(x,v\cond x',v')dxdv.
\end{align}
Then, integrating over $A_2$ that is around $[x,v] = f(x',v')$, we have
\begin{align}
    \int_{A_2} r(x,v\cond x',v')dxdv = & \int_{f(A_2)} dxdv\; \delta([x',v']-[x,v])
    \cdot \min\bigg\{\frac{p(f^{-1}(x,v))}{p(x',v')},  \bigg| \frac{\partial f(y)}{\partial y} \bigg|_{y=f^{-1}(x,v)} \bigg\} \bigg| \frac{\partial f^{-1}(x,v)}{\partial [x,v]} \bigg| = \\
    = & \int_{f(A_2)} dxdv\; \delta([x',v']-[x,v])
    \cdot \min\bigg\{\frac{p(f^{-1}(x,v))}{p(x',v')}\bigg|\frac{\partial f^{-1}(x,v)}{\partial [x,v]} \bigg|,  1 \bigg\}
\end{align}
Since $f$ is an involutive map, then $f^{-1} = f$, and $[x',v']$ lies in $f(A_2)$, where $A_2$ is an area around $[x,v] = f(x',v')$.
Thus, we have
\begin{align}
    \int_{A_2} t(x,v\cond x',v')dxdv = & \min \bigg\{1, \frac{p(f(x',v'))}{p(x',v')}\bigg| \frac{\partial f(x',v')}{\partial [x',v']} \bigg|\bigg\} = \int_{A_2} r(x,v\cond x',v')dxdv
\end{align}
Hence, $t(x',v'\cond x,v)$ satisfies the detailed balance in the joint space.
Moreover, that yields the detailed balance on the support of $p(x)$.
Indeed, reducing to the samples from $p(x)$, we have the transition kernel
\begin{align}
    \widehat{t}(x\cond x') = \int t(x,v\cond x',v')p(v'\cond x')dv'dv.
\end{align}
By definition, the reverse transition kernel is
\begin{align}
    \widehat{r}(x'\cond x) = \widehat{t}(x\cond x')\frac{p(x')}{p(x)} = \frac{p(x')}{p(x)} \int t(x,v\cond x',v')p(v'\cond x') dv'dv.
\end{align}
Since $t(x,v\cond x',v')$ satisfies the detailed balance, we have
\begin{align}
    \widehat{r}(x'\cond x) =& \frac{p(x')}{p(x)} \int t(x',v'\cond x,v)p(v'\cond x') \frac{p(x,v)}{p(x',v')} dv'dv = \int t(x',v'\cond x,v)p(v\cond x)dv'dv = \widehat{t}(x'\cond x)
\end{align}
Hence, $\widehat{t}(x\cond x')$ also satisfies the detailed balance.

\subsection{\cite{murray2012driving, neal2012view}}
\label{app:det_mcmc}

Here we formulate the algorithm from the papers \citep{murray2012driving, neal2012view}.
We consider one-dimensional target density $p(x)$ and some transition kernel $q(x'\cond x)$ that satisfy the fixed point equation with the target density.
For any kernel $q(x'\cond x)$ we can define the reverse transition kernel $r(x\cond x')$ in terms of so-called generalized detailed balance:
\begin{align}
    r(x\cond x')p(x') = q(x'\cond x)p(x).
\end{align}
Note that the reverse kernel is a correct distribution w.r.t. $x$, and also satisfy the fixed point equation:
\begin{align}
    \int dx\; r(x\cond x') = \frac{1}{p(x')} \int dx\; q(x'\cond x) p(x) = 1, \;\;\; \int dx'\; r(x\cond x')p(x') = \int dx'\; q(x'\cond x)p(x) = p(x).
\end{align}
Consider the joint distribution $p(x,u)=p(x)p(u)$, where $p(u) = \text{Uniform}[0,1]$.
For now, assume that at each iteration $u$ is sampled independently from the uniform distribution and the transition kernel is the deterministic function $f(x,v) = [x',v']$ defined as:
\begin{align}
    x' = F^{-1}_{q(\cdot\cond x)}(u), \;\;\;
    u' = F_{r(\cdot\cond x')}(x),
\end{align}
where $F_p$ is a CDF of a distribution with the density $p$.
To check the measure-preserving condition \eqref{eq:fpe}, we need to derive the determinant of the Jacobian of the $f$.
Using the chain rule, we have
\begin{align}
    \frac{\partial u'}{\partial u} = \frac{\partial u'}{\partial x'}\frac{\partial x'}{\partial u}, \;\;\; 
    \frac{\partial u'}{\partial x} = r(x\cond x') + \frac{\partial u'}{\partial x'}\frac{\partial x'}{\partial x}.
\end{align}
Then the Jacobian is
\begin{align}
    |J| = \bigg|\frac{\partial x'}{\partial x}\frac{\partial u'}{\partial u}-\frac{\partial x'}{\partial u}\frac{\partial u'}{\partial x}\bigg| = \frac{\partial x'}{\partial u} \bigg|\frac{\partial x'}{\partial x}\frac{\partial u'}{\partial x'} - \frac{\partial u'}{\partial x}\bigg| = \frac{r(x\cond x')}{q(x'\cond x)}.
\end{align}
Now, it is easy to check the measure preserving condition \eqref{eq:measure_preserve} using the definition of the reverse transition kernel.
\begin{align}
    p(f(x,u)) \bigg| \frac{\partial f(x,u)}{\partial [x,u]} \bigg| = p(x')p(u') \frac{r(x\cond x')}{q(x'\cond x)} = p(x) = p(x,u).
\end{align}

In the paper \citep{murray2012driving}, the authors propose to use some dependent random stream $d_t$ to update the auxiliary variable $u$ as $u_t = (u_{t-1} + d_t) \mod 1$, instead of sampling from the uniform.
In some cases, it is even possibly to eliminate all the stochasticity by letting $d_t$ be some constant irrational number: $d_t = c$.

\subsection{Proof of Trick \ref{th:moi} (Mixture of involutions)}
\label{app:moi_proof}

We remind that in the trick we consider the joint distribution $p(x,v,a) = p(x,v) p(a\cond x,v)$, and the family of involutions $f_a(x,v)$, i.e. $f_a(f_a(x,v)) = [x,v]$.
To make the calculations more concise, we denote the tuple $[x,v]$ as $y$.
Then the transition kernel for the distribution $p(y,a) = p(x,v,a)$ is
\begin{align}
\begin{split}
    t(y',a'\cond y,a) =& \delta([y',a']-[f_a(y),a]) \min\bigg\{1,\frac{p(f_a(y))p(a\cond f_a(y))}{p(y)p(a\cond y)} \bigg|\frac{\partial f_a(y)}{\partial y}\bigg|\bigg\} + \\
    &+ \delta([y',a']-[y,a]) \bigg(1-\min\bigg\{1,\frac{p(f_a(y))p(a\cond f_a(y))}{p(y)p(a\cond y)} \bigg|\frac{\partial f_a(y)}{\partial y}\bigg|\bigg\}\bigg).
\end{split}
\label{eq:moi_kernel}
\end{align}
Putting this transition kernel into the fixed point equation ($\int t(y',a'\cond y,a) p(y,a) dyda = p(y',a')$), we have
\begin{align}
\begin{split}
    &\int dyda \;\delta([y',a']-[f_a(y),a]) \min\bigg\{p(y,a), p(f_a(y))p(a\cond f_a(y)) \bigg|\frac{\partial f_a(y)}{\partial y}\bigg|\bigg\} + \\
    &+ p(y',a') - \int dyda\; \delta([y',a'] - [y,a])\min\bigg\{p(y,a),p(f_a(y))p(a\cond f_a(y)) \bigg|\frac{\partial f_a(y)}{\partial y}\bigg|\bigg\} = p(y',a').
\end{split}
\label{eq:fpe_for_moi}
\end{align}
From the last equation, we immediately obtain the equation
\begin{align}
    \min\bigg\{p(f_{a'}^{-1}(y'),a') \bigg|\frac{\partial f_{a'}^{-1}(y')}{\partial y'}\bigg|, p(y',a') \bigg\} =  \min\bigg\{p(y',a'),p(f_{a'}(y'),a') \bigg|\frac{\partial f_{a'}(y')}{\partial y'}\bigg|\bigg\},
\end{align}
which solutions in the space of $f_a$ include all involutive functions: $f_a(y) = f_a^{-1}(y)$.

To demonstrate that we must not change the variable $a$ let's try to apply some smooth function $g$ to propose a new $a$.
Then equation \eqref{eq:fpe_for_moi} becomes
\begin{align}
\begin{split}
    &\int dyda\; \delta([y',a']-[f_a(y),g(a)]) \min\bigg\{p(y,a), p(f_a(y), g(a)) \bigg|\frac{\partial f_a(y)}{\partial y}\bigg| \bigg|\frac{\partial g(a)}{\partial a}\bigg|\bigg\} + \\
    &+ p(y',a') - \int dyda \;\delta([y',a'] - [y,a])\min\bigg\{p(y,a),p(f_a(y), g(a)) \bigg|\frac{\partial f_a(y)}{\partial y}\bigg| \bigg|\frac{\partial g(a)}{\partial a}\bigg|\bigg\} = p(y',a'),
\end{split}
\end{align}
which yields the much stronger condition:
\begin{align}
\begin{split}
    &\int da\; \delta(a'-g(a)) \min\bigg\{p(f^{-1}_{a}(y'),a)\bigg|\frac{\partial f^{-1}_a(y')}{\partial y'}\bigg|, p(y', g(a))  \bigg|\frac{\partial g(a)}{\partial a}\bigg|\bigg\} =\\
    &= \min\bigg\{p(y',a'),p(f_{a'}(y'), g(a')) \bigg|\frac{\partial f_{a'}(y')}{\partial y'}\bigg| \bigg|\frac{\partial g(a')}{\partial a'}\bigg|\bigg\}
\end{split}\\
\begin{split}
    &\min\bigg\{p(f^{-1}_{g^{-1}(a')}(y'),g^{-1}(a'))\bigg|\frac{\partial f^{-1}_{g^{-1}(a')}(y')}{\partial y'}\bigg|\bigg|\frac{\partial g^{-1}(a')}{\partial a'}\bigg|, p(y', a')  \bigg\} =\\
    &= \min\bigg\{p(y',a'),p(f_{a'}(y'), g(a')) \bigg|\frac{\partial f_{a'}(y')}{\partial y'}\bigg| \bigg|\frac{\partial g(a')}{\partial a'}\bigg|\bigg\}
\end{split}
\end{align}

Looking for some solutions of this equation, we see that the involutivity of $g$ ($g(a) = g^{-1}(a)$) is not enough anymore.
Now, we also need $f^{-1}_{g^{-1}(a)}(y) = f_a(y)$.
By the assumption, $f_a$ is an involution; hence, we must guarantee $f_{g^{-1}(a)}(y) = f_a(y)$.
Thus, we end up with $g^{-1}(a) = g(a) = a$, what forces $g$ to be the identity mapping.
Actually, we can guarantee $f^{-1}_{g^{-1}(a)}(y) = f_a(y)$ with non-trivial $g$ if $f$ is not an involution.
We describe the latter in Trick \ref{th:direction}.

The detailed balance for kernel \eqref{eq:moi_kernel} follows directly from Proposition \ref{th:detailed_balance}, as well as the detailed balance for the collapsed kernel to the support of $p(y)$.
To bring more intuition here, one can consider the simple case of independent $a$: $p(a\cond y) = p(a)$, then the kernel $t(y'\cond y)$ can be considered as a linear mixture, where each kernel is reversible:
\begin{align}
\begin{split}
    t(y'\cond y) =& \int da\; p(a) \bigg[ \delta(y'-f_a(y)) \min\bigg\{1,\frac{p(f_a(y))}{p(y)} \bigg|\frac{\partial f_a(y)}{\partial y}\bigg|\bigg\} + \\
    &+ \delta(y'-y) \bigg(1-\min\bigg\{1,\frac{p(f_a(y))}{p(y)} \bigg|\frac{\partial f_a(y)}{\partial y}\bigg|\bigg\}\bigg) \bigg].
\end{split}
\end{align}
The general case $p(y,a) = p(a\cond y)p(y)$ is called state-depended mixture by \cite{geyer2003metropolis}.

\section{Special cases of Involutive MCMC}

\subsection{Metropolis-Hastings algorithm}
\label{app:MH}

\begin{algorithm}[H]
  \caption{The Metropolis-Hastings algorithm}
  \begin{algorithmic}  
    \INPUT{density of target distribution $\hat{p}(x) \propto p(x)$}
    \INPUT{proposal distribution $q(x'\cond x)$}
    \STATE initialize $x$
    \FOR{$i = 0\ldots n$}
        \STATE sample proposal point $x' \sim q(x'\cond x)$
        \STATE $P = \min\{1,\frac{\hat{p}(x')q(x\cond x')}{\hat{p}(x)q(x'\cond x)} \}$
        \STATE $ x_{i} =
            \begin{cases} 
            x' , \text{ with probability } P\\
            x , \text{ with probability } (1-P)
            \end{cases}$
        \STATE $x \gets x_{i}$
    \ENDFOR
    \OUTPUT{ $\{x_0,\ldots, x_n\}$}
  \end{algorithmic} 
  \label{alg:MH}
\end{algorithm}

To see that the MH algorithm is an instance of iMCMC, let's define the joint distribution as $p(x,v) = q(v\cond x) p(x)$ and the deterministic map as $f(x,v) = \begin{bmatrix} 0 & 1 \\ 1 & 0 \\ \end{bmatrix} \begin{bmatrix} x \\ v \end{bmatrix}$ (note that it is an involution).
For that case, we can write iMCMC transition kernel as
\begin{align}
    t(x',v'\cond x,v) = \; \delta ([x',v'] - [v,x])\min\bigg\{1,\frac{p(x',v')}{p(x,v)}\bigg\} + \delta([x',v']-[x,v]) \bigg(1 - \min\bigg\{1, \frac{p(v,x)}{p(x,v)}\bigg\}\bigg)
\end{align}
Then we substitute the last equation into the reduced transition kernel
\begin{align}
    \widehat{t}(x'\cond x) = \int\; dvdv' t(x',v'\cond x,v)q(v\cond x)
\end{align}
\begin{align}
    t(x'\cond x,v) =& \int dv'\; t(x',v'\cond x,v) = \delta(x'-v) \min\bigg\{1,\frac{p(x',x)}{p(x,v)}\bigg\} +  \delta(x'-x) \bigg(1 - \min\bigg\{1, \frac{p(v,x)}{p(x,v)}\bigg\}\bigg) \\
    \widehat{t}(x'\cond x) =& \int dv\; t(x'\cond x,v)q(v\cond x) = q(x'\cond x) \min\bigg\{1,\frac{p(x')q(x\cond x')}{p(x)q(x'\cond x)}\bigg\} + \\ & + \delta(x'-x) \int dv\; q(v \cond x) \bigg(1 - \min\bigg\{1, \frac{p(v)q(x\cond v)}{p(x)q(v\cond x)}\bigg\}\bigg) = q_{\text{MH}}(x'\cond x)
\end{align}
The last equation is the kernel of the conventional Metropolis-Hastings algorithm with proposal $q(x'\cond x)$.

Note that the following special cases can be obtained by the same involution $f(x,v) = \begin{bmatrix} 0 & 1 \\ 1 & 0 \\ \end{bmatrix} \begin{bmatrix} x \\ v \end{bmatrix}$ and different auxiliary distributions:
\begin{itemize}
    \item the Random-Walk Metropolis \citep{metropolis1953equation} (auxiliary $q(v\cond x) = q(x\cond v)$)
    \item  Metropolis-adjusted Langevin dynamics \citep{besag1994comments, roberts1998optimal}.
    \item Any kernel $q(v\cond x)$ that satisfy the detailed balance ($q(v\cond x)p(x)=q(x\cond v)p(v)$)
    \item Any independent sampler $p(x)$ (auxiliary $p(v)$).
\end{itemize}

\subsection{Mixture Proposal MCMC}
\label{app:MPMCMC}

\begin{algorithm}[H]
  \caption{Mixture Proposal MCMC}
  \begin{algorithmic}  
    \INPUT{density of target distribution $p(x)$}
    \INPUT{mixture proposal distribution $\int q_r(x'\cond a)q_f(a\cond x) da$}
    \STATE initialize $x$
    \FOR{$i = 0\ldots n$}
        \STATE sample $a \sim q_f(a\cond x)$
        \STATE sample $x' \sim q_r(x'\cond a)$
        \STATE $P = \min\bigg\{1, \frac{p(x')q_r(x\cond a)q_f(a\cond x')}{p(x)q_r(x'\cond a)q_f(a\cond x)}\bigg\}$
        \STATE $ x_{i} =
            \begin{cases} 
            x' , \text{ with probability } P\\
            x , \text{ with probability } (1-P)
            \end{cases}$
        \STATE $x \gets x_{i}$
    \ENDFOR
    \OUTPUT{ $\{x_0,\ldots, x_n\}$}
  \end{algorithmic} 
  \label{alg:MPMCMC}
\end{algorithm}
\vspace{-1em}
We formulate the algorithm from the paper \citep{habib2018auxiliary} in Algorithm \ref{alg:MPMCMC}.
To demonstrate that the iMCMC formalism includes this algorithm, we take the joint distribution of target variable $x$ and auxiliary variables $a,v$ as $p(x,a,v) = p(x)q_r(v\cond a)q_f(a\cond x)$.
The deterministic involution is $f(x,a,v) = [v,a,x]$.
Then the transition kernel in the joint space is
\begin{align}
    t(x',a',v'\cond x,a,v) = &\; \delta([x',a',v'] - [v,a,x]) \min\bigg\{1, \frac{p(x')q_r(v'\cond a')q_f(a'\cond x')}{p(x)q_r(v\cond a)q_f(a\cond x)}\bigg\} + \\ 
    &+ \delta([x',a',v'] - [x,a,v])\bigg(1-\min\bigg\{1, \frac{p(v)q_r(x\cond a)q_f(a\cond v)}{p(x)q_r(v\cond a)q_f(a\cond x)}\bigg\}\bigg).
\end{align}
This transitional kernel is equivalent to the Algorithm \ref{alg:MPMCMC}.
Indeed, the probability to accept the proposed state $v$ is the same as the acceptance probability in Algorithm \ref{alg:MPMCMC} and the state $v$ goes from the same proposal $\int da \;q_r(v\cond a) q(a\cond x)$.

To make the equivalence more apparent we derive formula (17) from \citep{habib2018auxiliary} by integrating the transition kernel $t(x',a',v'\cond x,a,v)$ over the corresponding coordinates.
That is
\begin{align}
    \widehat{t}(x',a'\cond x) & = \int dadv'dv\; t(x',a',v'\cond x,a,v) p(a,v\cond x) = \\
    & = q_r(x'\cond a')q_f(a'\cond x) \min\bigg\{1, \frac{p(x')q_r(x\cond a')q_f(a'\cond x')}{p(x)q_r(x'\cond a')q_f(a'\cond x)}\bigg\} + \\
    & + \delta(x'-x) q_f(a'\cond x) \bigg(1- \int dv \; q_r(v\cond a')\min\bigg\{1, \frac{p(v)q_r(x\cond a')q_f(a'\cond v)}{p(x)q_r(v\cond a')q_f(a'\cond x)}\bigg\}\bigg).
\end{align}

Note that if we further marginalize the kernel $\widehat{t}(x',a'\cond x)$ over $a'$ we obtain the kernel
\begin{align}
    \widehat{t}(x'\cond x) & = \int da' \; q_r(x'\cond a')q_f(a'\cond x) \min\bigg\{1, \frac{p(x')q_r(x\cond a')q_f(a'\cond x')}{p(x)q_r(x'\cond a')q_f(a'\cond x)}\bigg\} + \\
    & + \delta(x'-x) \bigg(1- \int dvda' \;  q_r(v\cond a')q_f(a'\cond x)\min\bigg\{1, \frac{p(v)q_r(x\cond a')q_f(a'\cond v)}{p(x)q_r(v\cond a')q_f(a'\cond x)}\bigg\}\bigg),
\end{align}
which is not equivalent to the Metropolis-Hastings kernel with the proposal 
\begin{align}
    \widetilde{q}(v\cond x) = \int da \;q_r(v\cond a) q(a\cond x).
\end{align}

\subsection{Multiple-Try Metropolis}
\label{app:MTM}

\begin{algorithm}[H]
  \caption{Multiple-Try Metropolis}
  \begin{algorithmic}  
    \INPUT{target density $p(x)$, proposal $q(y\cond x)$, nonnegative symmetric function $\lambda(x,y) = \lambda(y,x)$}
    \INPUT{denote weight function $w(x,y) = p(x)q(y\cond x)\lambda(x,y)$}
    \STATE initialize $x$
    \FOR{$i = 0\ldots n$}
        \STATE sample $y_1,\ldots,y_k \sim q(y_j \cond x)$
        \STATE evaluate weights $w_j = p(y_j)q(x\cond y_j)\lambda(y_j,x), \;\;\; j=1,\ldots,k$
        \STATE set $y = y_j$ with probability $w_j/(\sum_j w_j)$
        \STATE sample $x^*_1,\ldots,x^*_{k-1} \sim q(x_j \cond y)$
        \STATE set $x^*_k = x$
        \STATE $P = \min\bigg\{1,\frac{w(y_1,x)+\ldots +w(y_k,x)}{w(x^*_1,y)+\ldots +w(x^*_k,y)}\bigg\}$
        \STATE $ x_{i} =
            \begin{cases} 
            y , \text{ with probability } P\\
            x , \text{ with probability } (1-P)
            \end{cases}$
        \STATE $x \gets x_{i}$
    \ENDFOR
    \OUTPUT{samples $\{x_0,\ldots, x_n\}$}
  \end{algorithmic}
  \label{alg:MTM}
\end{algorithm}

We begin the proof with the recall of the Multiple-Try Metropolis (MTM) algorithm (Algorithm \ref{alg:MTM}).
To write MTM as Involutive MCMC, we consider the joint distribution and the family of involutions as follows.
\begin{align}
    p(x,y_1,\ldots,y_k,x^*_1,\ldots,x^*_{k-1},j) = p(x)\prod_{i=1}^{k} q(y_i\cond x) p(j\cond y_1,\ldots,y_k,x) \prod_{i=1}^{k-1} q(x^*_i\cond y_j), \\
    p(j\cond y_1,\ldots,y_k,x) = \frac{w(y_j,x)}{\sum_j w(y_j,x)}, \;\;\; w(x,y) = p(x)q(y\cond x)\lambda(x,y), \;\;\; j=1,\ldots,k\\
    f_j (x,y_1,\ldots,y_k,x^*_1,\ldots,x^*_{k-1},j) = [y_j,x^*_1,\ldots,x^*_{j-1},x,x^*_j,\ldots,x^*_{k-1},y_1,\ldots,y_j,y_{j-1},\ldots,y_{k},j]
\end{align}
That is, based on the value of the auxiliary variable $j\in \{1,\ldots,k\}$, we first swap $y_j$ and $x$, and then we swap the rest $(k-1)$ $y$'s with all of the $x^*$. 
Note that for the fixed $j$ that is an involution.
To check that iMCMC provides the equivalent chain, we evaluate the probability to accept $y_j$ as the next sample.
That is
\begin{align}
    P &= \min\bigg\{1,\frac{p(y_j)q(x\cond y_j)\prod_{i=1}^{k-1} q(x^*_i\cond y_j) p(j\cond x^*_1,\ldots,x^*_{j-1},x,x^*_j,\ldots,x^*_{k-1},y_j) \prod_{i=1,i\neq j}^{k} q(y_i\cond x)}{p(x)\prod_{i=1}^{k} q(y_i\cond x) p(j\cond y_1,\ldots,y_k,x) \prod_{i=1}^{k-1} q(x^*_i\cond y_j)}\bigg\} = \\
    &= \min\bigg\{1,\frac{p(y_j)q(x\cond y_j) p(j\cond x^*_1,\ldots,x^*_{j-1},x,x^*_j,\ldots,x^*_{k-1},y_j)}{p(x)q(y_j\cond x) p(j\cond y_1,\ldots,y_k,x)}\bigg\} =\\
    &= \min\bigg\{1,\frac{p(y_j)q(x\cond y_j) w(x,y_j) (\sum_{i=1}^k w(y_i,x))}{p(x)q(y_j\cond x) w(y_j,x) (\sum_{i=1}^{k-1}w(x^*_i,y_j) + w(x,y_j))}\bigg\} = \\
    &= \min\bigg\{1,\frac{p(y_j)q(x\cond y_j) p(x)q(y_j\cond x)\lambda(x,y_j) (\sum_{i=1}^k w(y_i,x))}{p(x)q(y_j\cond x) p(y_j)q(x\cond y_j)\lambda(y_j,x) (\sum_{i=1}^{k-1}w(x^*_i,y_j) + w(x,y_j))}\bigg\} = \\
    &= \min\bigg\{1,\frac{w(y_1,x)+\ldots +w(y_k,x)}{w(x^*_1,y)+\ldots + w(x^*_{k-1},y) +w(x,y)}\bigg\}.
\end{align}
Note that the distribution of $y$'s and $j$ is the same as in Algorithm \ref{alg:MTM}, hence, the probability to generate proposal $y_j$ is the same, as well as the probability to accept this proposal.

\subsection{Sample-Adaptive MCMC}
\label{app:SAMCMC}

\begin{algorithm}[H]
  \caption{Sample-Adaptive MCMC}
  \begin{algorithmic}  
    \INPUT{target density $p(x)$, integer $N$, aggregation function $g(x_1,\ldots,x_N)$, proposal $q\bigg(x_{N+1}\bigg| g(x_1,\ldots,x_N)\bigg)$}
    \STATE $\text{samples} = \emptyset$
    \STATE initialize set $S=\{x_1,\ldots,x_N\}$
    \FOR{$i = 0\ldots n$}
        \STATE sample $x_{N+1} \sim q\bigg(x_{N+1}\bigg| g(S)\bigg)$
        \STATE define $S_{-i} = (S \text{ with } x_i \text{ replaced with } x_{N+1}), \;\; S_{-(N+1)} = S$
        \STATE evaluate $\lambda_i = q\bigg(x_i\bigg| g(S_{-i})\bigg)/p(x_i), \;\;\; i = 1,\ldots,N+1$
        \STATE set $j = i$ with probability $\lambda_i/(\sum_{i=1}^{N+1}\lambda_i)$
        \STATE $S \gets S_{-j}$
        \STATE $\text{samples} = \text{samples} \cup S$
    \ENDFOR
    \OUTPUT{samples}
  \end{algorithmic}
  \label{alg:SAMCMC}
\end{algorithm}

We begin the proof with the recall of the Sample-Adaptive MCMC (SA-MCMC) algorithm (Algorithm \ref{alg:SAMCMC}).
In Algorithm \ref{alg:SAMCMC}, the output of function $g$ does not depend on the order of arguments, i.e. $g(x) = g(\pi(x))$, where $\pi$ is an arbitrary permutation of arguments.

To write SA-MCMC as Involutive MCMC, we consider the joint distribution and the family of involutions as follows.
\begin{align}
    p(x_1,\ldots,x_{N+1},j) = \prod_{i=1}^N p(x_i) q(x_{N+1}\cond g(x_1,\ldots,x_N)) p(j\cond x_1,\ldots,x_{N+1}), \\
    p(j\cond x_1,\ldots,x_{N+1}) = \frac{\lambda_j}{(\sum_{j=1}^{N+1}\lambda_j)}, \;\;\; \lambda_j = q(x_j\cond g(S_{-j}))/p(x_j), \;\;\; j=1,\ldots,N+1\\
    f_j(x_1,\ldots,x_{N+1},j) = f(x_1,\ldots,x_{j-1},x_{N+1},x_{j+1},\ldots,x_N,x_j,j)
\end{align}
Here $S_{-j}$ is the current set of samples $S = \{x_1,\ldots,x_N\}$, where $x_j$ is replaced with $x_{N+1}$, and $S_{-(N+1)} = S$. 
The involution family operates as follows. 
Based on the value of the auxiliary variable $j \in \{1,\ldots,N+1\}$, we swap $x_j$ and $x_{N+1}$ and leave the rest of arguments untouched.
For the fixed $j$, such function is an involution.
One more important thing to note is that now our target distribution is the product $\prod_{i=1}^N p(x_i)$.
To demonstrate that SA-MCMC is equivalent to Involutive MCMC with aforementioned distribution and involutions, we evaluate the probability to accept the point proposed by $f_j$.
\begin{align}
    P = \min\bigg\{1,\frac{p(x_{N+1})\prod_{i=1, i\neq j}^N p(x_i) q(x_j\cond g(S_{-j})) p(j\cond S_{-j},x_j)}{\prod_{i=1}^N p(x_i) q(x_{N+1}\cond g(S)) p(j\cond S,x_{N+1})}\bigg\} = \min\bigg\{1,\frac{p(x_{N+1}) q(x_j\cond g(S_{-j})) p(j\cond S_{-j},x_j)}{p(x_j) q(x_{N+1}\cond g(S)) p(j\cond S,x_{N+1})}\bigg\}
    \label{eq:samcmc_prob}
\end{align}

Now we define $S' = S_{-j}$ and $S'_{-i}\gets (S'$ with $i$-th element replaced by $x_j)$.
If we neglect the order of elements, then $S'_{-i} = S_{-i}$ for $i\neq j$, $S'_{-j} = S$ and $S'_{-(N+1)} = S_{-j}$.
Using the fact that the order of arguments in the aggregation function $g(\cdot)$ does not matter, we obtain
\begin{align}
    p(j\cond S_{-j},x_j) & = \frac{q(x_{N+1}\cond g(S))}{p(x_{N+1})\bigg(q(x_{N+1}\cond g(S))/p(x_{N+1}) + \sum_{i=1, i\neq j}^{N} q(x_{i}\cond g(S_{-i}))/p(x_{i}) + q(x_j\cond g(S_{-j}))/p(x_j) \bigg)} \\
    & = \frac{q(x_{N+1}\cond g(S))}{p(x_{N+1})\bigg(\sum_{i=1}^{N+1} q(x_{i}\cond g(S_{-i}))/p(x_{i})\bigg)}
\end{align}

Putting this equation into \eqref{eq:samcmc_prob}, we obtain
\begin{align}
    P &= \min\bigg\{1,\frac{q(x_j\cond g(S_{-j}))}{p(x_j) p(j\cond S,x_{N+1})\bigg(\sum_{i=1}^{N+1} q(x_{i}\cond g(S_{-i}))/p(x_{i})\bigg)}\bigg\} = 1.
\end{align}
Thus, generating the auxiliary variable $j$ we accept the point $f_j(x_1,\ldots,x_{N+1},j)$ with probability $1$.
Since the distribution of $j$ and the corresponding point $f_j(x_1,\ldots,x_{N+1},j)$ are the same as in Algorithm \ref{alg:SAMCMC}, we have obtained the equivalent scheme in terms of Involutive MCMC.

\subsubsection{Generalization of Sample-Adaptive MCMC}
\label{app:gSAMCMC}

From the equations above it is easy to discard the permutation-invariance property of $g(\ldots)$.
Then we just denote $S$ to be an ordered array $S = [x_1,\ldots, x_N]$ instead of a set, and accept the proposed swap with probability
\begin{align}
    P_j = \min\bigg\{1,\frac{p(x_{N+1}) q(x_j\cond S_{-j}) p(j\cond S_{-j},x_j)}{p(x_j) q(x_{N+1}\cond S) p(j\cond S,x_{N+1})}\bigg\}.
\end{align}
Then the pseudo-code of the algorithm slightly changes (see Algorithm \ref{alg:gSAMCMC}).

\begin{algorithm}[H]
  \caption{Generalized Sample-Adaptive MCMC}
  \begin{algorithmic}  
    \INPUT{target density $p(x)$, integer $N$, proposal $q\bigg(x_{N+1}\bigg| x_1,\ldots,x_N\bigg)$}
    \STATE $\text{samples} = \emptyset$
    \STATE initialize array $S=[x_1,\ldots,x_N]$
    \FOR{$i = 0\ldots n$}
        \STATE sample $x_{N+1} \sim q\bigg(x_{N+1}\bigg| S\bigg)$
        \STATE define $S_{-i} = S (\text{ with } x_i \text{ replaced by } x_{N+1}), \;\; S_{-(N+1)} = S$
        \STATE evaluate $\lambda_i = q\bigg(x_i\bigg| S_{-i}\bigg)/p(x_i), \;\;\; i = 1,\ldots,N+1$
        \STATE set $j = i$ with probability $\lambda_i/(\sum_{i=1}^{N+1}\lambda_i)$
        \STATE evaluate acceptance probability $P = \min\bigg\{1,\frac{p(x_{N+1}) q(x_j\cond S_{-j}) p(j\cond S_{-j},x_j)}{p(x_j) q(x_{N+1}\cond S) p(j\cond S,x_{N+1})}\bigg\}$
        \STATE $S \gets \begin{cases} S_{-j}, \text{ with probability } P\\
        S, \text{ with probability } (1-P)
        \end{cases}$
        \STATE $\text{samples} = \text{samples} \cup S$
    \ENDFOR
    \OUTPUT{samples}
  \end{algorithmic}
  \label{alg:gSAMCMC}
\end{algorithm}

\subsection{Reversible-Jump MCMC}
\label{app:RJMCMC}

\subsubsection{Reversible-Jump MCMC from \citep{green2009reversible}}

\begin{algorithm}[H]
  \caption{Reversible-Jump MCMC from \citep{green2009reversible}}
  \begin{algorithmic}
    \INPUT{target density $p(x^{(k)},k)$, auxiliary distributions $q(u\cond m)$ and $q'(u\cond m)$, move functions $h_m(x,u)$}
    \STATE initialize $\text{state} = [x^{(k)}, k]$
    \FOR{$i = 0\ldots n$}
        \STATE unpack $[x^{(k)}, k] \gets \text{state}$
        \STATE sample move type $m \sim p(m\cond x^{(k)},k)$
        \STATE sample auxiliary $u \sim q(u\cond m)$
        \STATE move type $m$ defines $k'$
        \STATE evaluate $[x^{(k')}, u'] = h_{m}(x^{(k)}, u)$
        \STATE evaluate $P = \min\bigg\{1, \frac{p(x^{(k')},k') p(m\cond x^{(k')},k') q'(u'\cond m)}{p(x^{(k)},k) p(m\cond x^{(k)},k) q(u\cond m)}\bigg| \frac{\partial h_{m}}{\partial [x^{(k)}, u]}\bigg|\bigg\}$
        \STATE accept $\text{state} \gets \begin{cases} [x^{(k')}, k'], \text{ with probability } P\\
        [x^{(k)}, k], \text{ with probability } (1-P)
        \end{cases}$
        \STATE $\text{state}_i \gets \text{state}$
    \ENDFOR
    \OUTPUT{samples $\{\text{state}_0,\ldots,\text{state}_n\}$}
  \end{algorithmic}
  \label{alg:RJMCMC}
\end{algorithm}

Reversible-Jump MCMC \citep{green1995reversible} has multiple formulations, which vary significantly both in notation used and in the sampling procedure.
Here we choose to stay close to \citep{green2009reversible} for illustrative purposes (see pseudo-code in Algorithm \ref{alg:RJMCMC}).
Note that the move type $m$ index both models $k$ and $k'$, as well as the smooth map $h_m$.
Indeed, for a proper scheme, auxiliary distributions $q'(u\cond m)$ and $q(u\cond m)$ are defined such that the dimension of $[x^{(k)}, u]$ matches the dimension of $[x^{(k')}, u']$ and the dimension for the input of $h_m$.

To describe Algorithm \ref{alg:RJMCMC} in terms of iMCMC, we consider the joint distribution:
\begin{align}
    p(x,k,m,u) = p(x^{(k)},k) p(m\cond x^{(k)},k) p(u\cond m,k),
\end{align}
where we define $p(u\cond m,k)$ such that for the move type $m$ that goes from $k$ to $k'$ we have $p(u\cond m,k) = q(u\cond m)$ and $p(u\cond m,k') = q'(u\cond m)$.
We can do it because $m$ defines both models $k$ and $k'$.
The family of involutions is then defined as follows.
\begin{align}
    f_m(x^{(k)},u,k) = [h_m(x^{(k)},u),k'] = [x^{(k')},u',k'], \;\;\; f_m(x^{(k')},u',k') = [h_m^{-1}(x^{(k')},u'),k] = [x^{(k)},u,k]
\end{align}
Here index $m$ choose such involution that map model index $k$ to $k'$ and vice versa.
As well as in \citep{green2009reversible}, mapping from $k'$ to $k$ we apply the inverse $h^{-1}_m$.
For a concrete example of move types and functions $h_m$, we refer the reader to Section 3 of \citep{green2009reversible}.
The acceptance probability then is in total agreement with Algorithm \ref{alg:RJMCMC}:
\begin{align}
    P = \min\bigg\{1, \frac{p(x^{(k')},k') p(m\cond x^{(k')},k') q'(u'\cond m)}{p(x^{(k)},k) p(m\cond x^{(k)},k) q(u\cond m)}\bigg| \frac{\partial h_{m}}{\partial [x^{(k)}, u]}\bigg|\bigg\}.
\end{align}

\subsubsection{Another formulation}

In the previous section, we encapsulate the knowledge about the next proposed model in the index $m$.
However, the formulation becomes more transparent if we sample the index of the next proposed model explicitly.
The following algorithm can be seen as a more general version of the formulation of Reversible-Jump MCMC from \citep{gagnon2019non}.
That is, consider the joint distribution
\begin{align}
    p(x,k,j,u) = p(x^{(k)},k) p(j\cond x^{(k)},k) p(u^{(k)}\cond x^{(k)},k,j),
\end{align}
where $j$ is the index of the next model.
Here we add superscripts for $u$ to highlight that the choice of auxiliary variables relies on the current model $k$.
Usually this is done such that all vectors lie in the same vector space, i.e. $[x^{(k)}, u^{(k)}] \in \mathbb{R}^d\;\; \forall k$.
The involution $f$ then is
\begin{align}
    f(x^{(k)},u^{(k)},k,j) = [h_{kj}(x^{(k)},u^{(k)}), j, k] = [x^{(j)},u^{(j)},j,k], \;\;\; h_{jk}(x^{(j)},u^{(j)}) = h^{-1}_{kj}(x^{(j)},u^{(j)}) = [x^{(k)},u^{(k)}].
\end{align}
Here the involution $f$ maps $[x,u]$ based on the indeces of the current model $k$ and the next model $j$.
Note that mapping from $k$ to $j$ via $h_{kj}$ we are obliged to perform the inverse map $h_{jk}$ using the inverse function $h_{kj}^{-1}$.
The acceptance probability is then
\begin{align}
    P = \min\bigg\{1, \frac{p(x^{(j)},j) p(k\cond x^{(j)},j) p(u^{(j)}\cond x^{(j)},j,k)}{p(x^{(k)},k) p(j\cond x^{(k)},k) p(u^{(k)}\cond x^{(k)},k,j)}\bigg| \frac{\partial h_{kj}}{\partial [x^{(k)}, u^{(k)}]}\bigg|\bigg\}.
\end{align}
See the pseudo-code in Algorithm \ref{alg:gRJMCMC}.
Note that unlike Algorithm \ref{alg:RJMCMC}, here we have a single smooth map from model $k$ to model $j$.
This limitation can be easily removed via Trick \ref{th:moi} by considering the family of involutions
\begin{align}
    f_m(x^{(k)},u^{(k)},k,j) = [h_{mkj}(x^{(k)},u^{(k)}), j, k] = [x^{(j)},u^{(j)},j,k], \;\;\; h_{mjk}(x^{(j)},u^{(j)}) = h^{-1}_{mkj}(x^{(j)},u^{(j)}) = [x^{(k)},u^{(k)}],
\end{align}
where we can sample index $m$ conditioned on the current state $[x^{(k)},u^{(k)},k,j]$.

Finally, we discuss the usage of Tricks from Section \ref{sec:tricks} here.
Trick \ref{th:moi} is explicitly used here when we define a family of involutions and stochastically choose one from the family.
The auxiliary direction from Trick \ref{th:direction} here is in the form of indices $k$ and $j$, which define the smooth map $h_{kj}$ and its inverse $h_{jk} = h_{kj}^{-1}$.
Trick \ref{th:mod} can be found here if we define the target distribution as
\begin{align}
    p(x^{(k)},u^{(k)},k) = p(x^{(k)},k) p(u^{(k)}\cond x^{(k)},k),
\end{align}
in order to match the dimensions of all models $[x^{(k)}, u^{(k)}] \in \mathbb{R}^d\;\; \forall k$.
As well as in Trick \ref{th:mod}, we sample from extended distribution $p(x^{(k)},u^{(k)},k)$, and then discard all $u^{(k)}$.

\begin{algorithm}[H]
  \caption{Reversible-Jump MCMC}
  \begin{algorithmic}
    \INPUT{target density $p(x^{(k)},k)$, distribution of next models $p(j\cond x^{(k)},k)$, auxiliary distributions $p(u^{(k)}\cond x^{(k)},k,j)$}
    \STATE initialize $\text{state} = [x^{(k)}, k]$
    \FOR{$i = 0\ldots n$}
        \STATE unpack $[x^{(k)}, k] \gets \text{state}$
        \STATE sample next model $j \sim p(j\cond x^{(k)},k)$
        \STATE sample auxiliary $u^{(k)} \sim p(u^{(k)}\cond x^{(k)},k,j)$
        \STATE propose $[x^{(j)}, u^{(j)}] = h_{kj}(x^{(k)}, u^{(k)})$
        \STATE evaluate $P = \min\bigg\{1, \frac{p(x^{(j)},j) p(k\cond x^{(j)},j) p(u^{(j)}\cond x^{(j)},j,k)}{p(x^{(k)},k) p(j\cond x^{(k)},k) p(u^{(k)}\cond x^{(k)},k,j)}\bigg| \frac{\partial h_{kj}}{\partial [x^{(k)}, u^{(k)}]}\bigg|\bigg\}$
        \STATE accept $\text{state} \gets \begin{cases} [x^{(j)}, j], \text{ with probability } P\\
        [x^{(k)}, k], \text{ with probability } (1-P)
        \end{cases}$
        \STATE $\text{state}_i \gets \text{state}$
    \ENDFOR
    \OUTPUT{samples $\{\text{state}_0,\ldots,\text{state}_n\}$}
  \end{algorithmic}
  \label{alg:gRJMCMC}
\end{algorithm}

\subsection{Hybrid Monte Carlo}
\label{app:HMC}

\begin{algorithm}[H]
  \caption{Hybrid Monte Carlo}
  \begin{algorithmic}  
    \INPUT{joint density $p(x,v)=p(x)p(v)$, auxiliary distribution $p(v) = \Normal(v\cond 0,1)$, number of Leap-Frog steps $k$, step size $\eps$}
    \STATE initialize $x$
    \FOR{$i = 0\ldots n$}
        \STATE sample $v \sim \Normal(v\cond 0,1)$
        \STATE propose $[x',v'] = FL^k(x,v)$
        \STATE evaluate $P = \min\{1, \frac{p(x',v')}{p(x,v)}\}$
        \STATE accept $x \gets \begin{cases}
        x', \text{ with probability } P\\
        x, \text{ with probability } (1-P)\\
        \end{cases}$
        \STATE $x_i \gets x$
    \ENDFOR
    \OUTPUT{$\{x_0,\ldots, x_n\}$}
  \end{algorithmic} 
  \label{alg:hmc}
\end{algorithm}

Hybrid Monte Carlo \citep{duane1987hybrid} relies on the numerical integration of Hamiltonian dynamics via the Leap-Frog operator $L$.
For target density $p(x)$, the Hamiltonian is defined as $H(x,v) = -\log p(x,v)$, where $p(x,v)=p(x)p(v)$ is the joint distribution, and $p(v)=\Normal(v\cond 0,1)$ is the auxiliary distribution.
In the case of independent $v$ (i.e., $p(x,v)=p(x)p(v)$), the Leap-Frog operator $L: [x(t),v(t)] \to [x(t+\eps),v(t+\eps)]$ is defined as follows.
\begin{align}
\label{eq:lf_1}
    v(t+\eps/2) =& v(t) - \frac{\eps}{2} \nabla_x (-\log p(x(t))) \\
    x(t+\eps) =& x(t) + \eps \nabla_v (-\log p(v(t+\eps/2))) \\
    v(t+\eps) =& v(t+\eps/2) - \frac{\eps}{2} \nabla_x (-\log p(x(t+\eps))) 
\end{align}
Flip operator $F$ denotes the negation of the auxiliary variable (momentum) $v$: $F:[x,v]\to [x,-v]$.
These operators together yields the involutive map $FL$, which is used in Algorithm \ref{alg:hmc}.
To demonstrate this, we demonstrate that $FLFL = 1$, i.e. double application of the operator $FL$ results in identity function.
\begin{align}
    v(t+\eps/2) = & v(t) - \frac{\eps}{2} \nabla_x (-\log p(x(t))) \\
    x(t+\eps) = & x(t) + \eps \nabla_v (-\log p(v(t+\eps/2))) \\
    v(t+\eps) = & v(t+\eps/2) - \frac{\eps}{2} \nabla_x (-\log p(x(t+\eps)))\\
    v(t+3/2\eps) = & -v(t+\eps) - \frac{\eps}{2} \nabla_x (-\log p(x(t+\eps)))= -v(t+\eps/2) \\
    x(t+2\eps) = & x(t+\eps) + \eps \nabla_v (-\log p(v(t+3/2\eps))) = x(t) \\
    v(t+2\eps) = & v(t+3/2\eps) - \frac{\eps}{2} \nabla_x (-\log p(x(t+2\eps))) = -v(t)
\end{align}
Note that here we greatly rely on the symmetry $p(v)=p(-v)$.
After the last equation we negate the momentum variable once again yielding $FLFL: [x(t), v(t)] \to [x(t), v(t)]$.
Note that having $FLFL = 1$ we can easily obtain the inverse of the Leap-Frog operator $L^{-1} = FLF$.
Using the formula for the inverse Leap-Frog we have 
\begin{align}
    FL^kFL^k = FL^kFLFFL^{k-1} = FL^{k-1}FL^{k-1} = \ldots = FLFL = 1.
\end{align}
Thus, an arbitrary number of $L$ can be composed in the involution $FL^k$.

Using the involution $FL^k$, the formulation of HMC in terms of iMCMC is now straightforward.
Consider the joint distribution $p(x,v) = p(x)\Normal(v\cond 0,1)$ and the involutive function $FL^k$, the acceptance probability according to iMCMC (Algorithm \ref{alg:imcmc}) is then
\begin{align}
    P = \min\bigg\{1,\frac{p(FL^k(x,v))}{p(x,v)}\bigg|\frac{\partial FL^k}{\partial [x,v]}\bigg|\bigg\}.
\end{align}
Finally, it is easy to see that $FL^k$ is volume-preserving since the transformations on the each step of $L$ are volume-preserving, e.g. \eqref{eq:lf_1} maps $[x(t),v(t)] \to [x(t),v(t+\eps/2)]$ since it is an identity map w.r.t. $x(t)$, and $|\partial v(t+\eps/2)/ \partial v(t)| = 1$ it is volume-preserving.

Another possible way to represent HMC in terms of iMCMC is to use Trick \ref{th:direction} and introduce the directional variable $p(d) = \text{Uniform}\{-1,+1\}$.
Then the involutive map is defined as
\begin{align}
    f(x,v,d) = [T_d(x,v),-d], \;\;\; T_{d=+1} = L, \;\;\; T_{d=-1} = L^{-1}.
\end{align}
This formulation allow for a more general formulation that does not rely on the symmetry $p(v)=p(-v)$ as HMC.
Indeed, the inverse Leap-Frog operator $L^{-1}$ can be obtained just by the inversion of the time:
\begin{align}
    v(t-\eps/2) =& v(t) + \frac{\eps}{2} \nabla_x (-\log p(x(t))) \\
    x(t-\eps) =& x(t) - \eps \nabla_v (-\log p(v(t-\eps/2))) \\
    v(t-\eps) =& v(t-\eps/2) + \frac{\eps}{2} \nabla_x (-\log p(x(t-\eps))) 
\end{align}

\subsection{RMHMC}
\label{app:RMHMC}

\begin{algorithm}[H]
  \caption{Riemann Manifold HMC}
  \begin{algorithmic}  
    \INPUT{joint density $p(x,v)$, auxiliary distribution $p(v) = \Normal(v\cond 0,G(x))$, number of Leap-Frog steps $k$, step size $\eps$}
    \STATE initialize $x$
    \FOR{$i = 0\ldots n$}
        \STATE sample $v \sim \Normal(v\cond 0,G(x))$
        \STATE propose $[x',v'] = FL^k(x,v)$
        \STATE evaluate $P = \min\{1, \frac{p(x',v')}{p(x,v)}\}$
        \STATE accept $x \gets \begin{cases}
        x', \text{ with probability } P\\
        x, \text{ with probability } (1-P)\\
        \end{cases}$
        \STATE $x_i \gets x$
    \ENDFOR
    \OUTPUT{$\{x_0,\ldots, x_n\}$}
  \end{algorithmic} 
  \label{alg:rmhmc}
\end{algorithm}

In Riemann Manifold HMC \citep{girolami2011riemann}, the authors propose to take into account the ``curvature'' of the space during sampling by considering the following Hamiltonian
\begin{equation}
    H(x, v) = - \log p(x) + \tfrac{1}{2}\log|G(x)| + \tfrac{1}{2}v^TG(x)^{-1}v.
\end{equation}
As you can see from Algorithm \ref{alg:rmhmc}, the pseudo-code for RMHMC is almost the same as for HMC (see \ref{app:HMC}).
The key difference between them is the integration operator $L$.
Since the Hamiltonian $H(x,v)$ is not separable, we need to use the implicit numerical scheme to guarantee volume-preserving and involutive properties.
The integration operator $L$ is defined as follows.
\begin{align}
\label{eq:implicit_lf_1}
    &v(t+\eps/2) = v(t) - \frac{\eps}{2}\nabla_{x}H(x(t), v(t+\eps/2))\\
\label{eq:implicit_lf_2}
    &x(t+\eps/2) = x(t) + \frac{\eps}{2}\nabla_{v}H(x(t), v(t+\eps/2))\\
\label{eq:implicit_lf_3}
    &x(t+\eps) = x(t+\eps/2) + \frac{\eps}{2}\nabla_{v}H(x(t+\eps),  v(t+\eps/2))\\
\label{eq:implicit_lf_4}
    &v(t+\eps) = v(t+\eps/2) - \frac{\eps}{2}\nabla_{x}H(x(t+\eps),  v(t+\eps/2))
\end{align}
The involution can be constructed as $FL$, where $F$ is the negation of $v$: $F: [x,v] \to [x,-v]$.
To demonstrate this, we integrate further in time from $[x(t+\eps), -v(t+\eps)]$ obtaining $FLFL = 1$ (double application yields identity function).
That is, applying step \eqref{eq:implicit_lf_1}, we get
\begin{align}
    & v(t+3/2\eps) = -v(t+\eps) - \frac{\eps}{2}\nabla_{x}H(x(t+\eps), v(t+3/2\eps))\\
    & v(t+\eps) = -v(t+3/2\eps) - \frac{\eps}{2}\nabla_{x}H(x(t+\eps), -v(t+3/2\eps)) \implies -v(t+3/2\eps) = v(t+\eps/2)\\
\end{align}
Here we use $\nabla_x H(x,v) = \nabla_x H(x,-v)$.
Further, applying step \eqref{eq:implicit_lf_2}, we get
\begin{align}
    x(t+3/2\eps) = x(t+\eps) + \frac{\eps}{2}\nabla_{v}H(x(t+\eps), v(t+3/2\eps)) = x(t+\eps/2),
\end{align}
where we use $\nabla_v H(x,-v) = -\nabla_v H(x,v)$.
The last two steps \eqref{eq:implicit_lf_3} and \eqref{eq:implicit_lf_4} follow the same logic.
\begin{align}
    &x(t+2\eps) = x(t+3/2\eps) + \frac{\eps}{2}\nabla_{v}H(x(t+2\eps),  v(t+3/2\eps)) = x(t)\\
    &v(t+2\eps) = v(t+3/2\eps) - \frac{\eps}{2}\nabla_{x}H(x(t+2\eps),  v(t+3/2\eps)) = -v(t)
\end{align}
Further negation of $-v(t)$ results in the initial point $[x(t),v(t)]$.
Thus, $FL$ is an involution ($FLFL=1$) and $FL^k$ is also an involution:
\begin{align}
    FL^{k}FL^{k} = FL^{k-1}F(FLFL)L^{k-1}= FL^{k-1}FL^{k-1} = \ldots = 1.
\end{align}

Using the involution $FL^k$, the formulation of RMHMC in terms of iMCMC is now straightforward.
Consider the joint distribution $p(x,v) = p(x)\Normal(v\cond 0,G(x))$ and the involutive function $FL^k$, the acceptance probability according to iMCMC (Algorithm \ref{alg:imcmc}) is then
\begin{align}
    P = \min\bigg\{1,\frac{p(FL^k(x,v))}{p(x,v)}\bigg|\frac{\partial FL^k}{\partial [x,v]}\bigg|\bigg\}.
\end{align}
Finally, it is easy to see that $FL^k$ is volume-preserving.
For illustrative purposes, we evaluate the Jacobian of the first two steps \eqref{eq:implicit_lf_1} and \eqref{eq:implicit_lf_2}.
\begin{align}
    & \frac{\partial x(t+\eps/2)}{\partial x(t)} = 1 + \frac{\eps}{2}\nabla_{vx} H(x(t),v(t+\eps/2)) + \frac{\eps}{2} \nabla_{vv} H(x(t),v(t+\eps/2)) \frac{\partial v(t+\eps/2)}{\partial x(t)} \\ 
    & \frac{\partial x(t+\eps/2)}{\partial v(t)} =   \frac{\eps}{2} \nabla_{vv} H(x(t),v(t+\eps/2)) \frac{\partial v(t+\eps/2)}{\partial v(t)} \\
    & \frac{\partial v(t+\eps/2)}{\partial x(t)} = -  \frac{\eps}{2} \nabla_{xx} H(x(t),v(t+\eps/2)) \\
    & \frac{\partial v(t+\eps/2)}{\partial v(t)} = 1 -  \frac{\eps}{2} \nabla_{xv} H(x(t),v(t+\eps/2)) \frac{\partial v(t+\eps/2)}{\partial v(t)} 
\end{align}
\begin{align}
    \bigg|\frac{\partial FL^k}{\partial [x,v]}\bigg| = \bigg(1 + \frac{\eps}{2}\nabla_{vx} H(x(t),v(t+\eps/2))\bigg)\frac{\partial v(t+\eps/2)}{\partial v(t)} = 1
\end{align}

\subsection{NeuTra}
\label{app:neutra}

\begin{algorithm}[H]
  \caption{NeuTra}
  \begin{algorithmic}
    \INPUT{target density $p_x(x)$, auxiliary density $p(v)=\Normal(v\cond 0, 1)$, flow $T(x)$}
    \STATE initialize $z$
    \FOR{$i = 0\ldots n$}
    \STATE sample $v \sim p(v)=\Normal(v\cond 0, 1)$
    \STATE propose $[z',v'] = FL^k(z,v)$, where the target density for Leap-Frog is $p_z(z,v)=p_x(T(z))|\frac{\partial T}{\partial z}|p(v)$
    \STATE evaluate $P = \min\bigg\{1, \frac{p_z(z',v')}{p_z(z,v)}\bigg\}$
    \STATE accept $x \gets \begin{cases} z', \text{ with probability } P\\
    z, \text{ with probability } (1-P)
    \end{cases}$
    \STATE $x_i \gets T(z)$
    \ENDFOR
    \OUTPUT{samples $\{x_0,\ldots,x_n\}$}
  \end{algorithmic}
  \label{alg:neutra}
\end{algorithm}

In the recent paper \citep{hoffman2019neutra}, the authors learn an invertible transformation $T^{-1}: X \to Z$ to map the target random variable $x \in X$ with the density $p_x(x)$ into another random variable $z \in Z$, which has more simple geometry of density levels.
Further, they run HMC in $Z$ with the target density $p_z(z) = p_x(T(z))|\partial T/\partial z|$.
Finally, one can obtain samples in the original space $X$ by mapping the collected samples using $T: Z \to X$.
We provide the pseudo-code in Algorithm \ref{alg:neutra}.

A straightforward application of Trick \ref{th:embed_involution} allows for iMCMC formulation of NeuTra.
That is, the joint distribution is just the same as in HMC
\begin{align}
    p(x,v) = p_x(x)\Normal(v\cond 0,1).
\end{align}
For the involutive map, we take
\begin{align}
    f(x,v) = \begin{bmatrix} T \\ 1 \end{bmatrix} \circ F \circ L^k \circ \begin{bmatrix} T^{-1} \\ 1 \end{bmatrix} \begin{bmatrix} x \\ v \end{bmatrix},
\end{align}
where $F$ is the velocity flip operator, $L$ is the Leap-Frog, and the notation $\begin{bmatrix} T^{-1} \\ 1 \end{bmatrix} \begin{bmatrix} x \\ v \end{bmatrix}$ means element-wise application $(x,v) \to (T^{-1}(x), v)$.
Note that the only necessary condition for the operators $L$ and $F$ is the $(F\circ L^k)^{-1} = F\circ L^k$.
Then, by the straightforward evaluation $f(f(x,v))$ we can see that $f$ is an involution. 
To obtain an equivalent sampler to NeuTra we choose the joint density for $L$ as $p(z,v) = p_x(T(z))|\partial T/\partial z| p(v)$.
Thus, we obtain the same dynamics in $Z$.
However, note that iMCMC assumes the acceptance test in the original space $X$, while NeuTra performs the acceptance test in $Z$.
Nevertheless, for an initial point $x$ and the velocity $v \sim p(v)$, Algorithm \ref{alg:imcmc} gives us the following acceptance test
\begin{align}
    P = \min\bigg\{1, \frac{p(f(x,v))}{p(x,v)} \bigg| \frac{\partial f(x,v)}{\partial [x,v]} \bigg| \bigg\}, \;\;\; f(x,v) = \begin{bmatrix} T \\ 1 \end{bmatrix} \circ F \circ L^k \circ \begin{bmatrix} T^{-1} \\ 1 \end{bmatrix} \begin{bmatrix} x \\ v \end{bmatrix}
\end{align}
Using the chain rule, we have
\begin{align}
    \bigg| \frac{\partial f(x,v)}{\partial [x,v]} \bigg| = \bigg| \frac{\partial T}{\partial y} \bigg|_{y=FL^kT^{-1}(x)} \bigg| \frac{\partial T^{-1}}{\partial x} \bigg| = \bigg| \frac{\partial T}{\partial y} \bigg|_{y=FL^kT^{-1}(x)} \bigg| \frac{\partial T}{\partial y} \bigg|^{-1}_{y=T^{-1}(x)} 
\end{align}
Denoting $z = T^{-1}(x)$, and $[z',v'] = FL^k(z,v)$, we have
\begin{align}
    P &= \min\bigg\{1, \frac{p_x(T(z'))p(v')}{p_x(T(z))p(v)} \bigg| \frac{\partial T}{\partial y} \bigg|_{y=z'} \bigg| \frac{\partial T}{\partial y} \bigg|^{-1}_{y=z} \bigg\} = \min\bigg\{1, \frac{p_z(z',v')}{p_z(z,v)}\bigg\}.
\end{align}
Thus, we obtain the same acceptance probability, and, hence, equivalent kernel to Algorithm \ref{alg:neutra}.

\subsection{A-NICE-MC}
\label{app:nicemc}

\begin{algorithm}[H]
  \caption{A-NICE-MC}
  \begin{algorithmic}
    \INPUT{target density $p(x,v) = p(x)\Normal(v\cond 0,1)$, NICE-proposal $T(x,v)$ and $T^{-1}(x,v)$}
    \STATE initialize $x$
    \FOR{$i = 0\ldots n$}
    \STATE sample $v \sim \Normal(v\cond 0, 1)$
    \STATE sample $d \sim \text{Uniform}\{-1,+1\}$
    \STATE propose $[x',v'] = T_{d}(x,v)$, where $T_{d=+1}=T$ and $T_{d=-1}=T^{-1}$
    \STATE evaluate $P = \min\bigg\{1, \frac{p(x',v')}{p(x,v)}\bigg\}$
    \STATE accept $x \gets \begin{cases} x', \text{ with probability } P\\
    x, \text{ with probability } (1-P)
    \end{cases}$
    \STATE $x_i \gets x$
    \ENDFOR
    \OUTPUT{samples $\{x_0,\ldots,x_n\}$}
  \end{algorithmic}
  \label{alg:nicemc}
\end{algorithm}

We recall A-NICE-MC \citep{song2017nice} in Algorithm \ref{alg:nicemc}.
The core part of the algorithm is the volume-preserving NICE proposal $T(x,v)$, which is learned before the sampling.
Trick \ref{th:direction} with directional variable $d$ allows for a straightforward formulation of A-NICE-MC in terms of iMCMC.
Consider the joint distribution
\begin{align}
    p(x,v,d) = p(x)\Normal(v\cond 0,1)p(d), \;\;\; p(d) = \text{Uniform}\{-1,+1\},
\end{align}
and the involution
\begin{align}
    f(x,v,d) = [T_{d}(x,v),-d], \;\;\; T_{d=+1}=T,\;\; T_{d=-1}=T^{-1}.
\end{align}
Then it is easy to see that the acceptance probability of iMCMC (Algorithm \ref{alg:imcmc}) is the same as the probability $P$ in Algorithm \ref{alg:nicemc}.

\subsection{L2HMC}
\label{app:l2hmc}

\begin{algorithm}[H]
  \caption{L2HMC}
  \begin{algorithmic}
    \INPUT{target density $p(x,v) = p(x)\Normal(v\cond 0,1)$, proposal $T(x,v)$ and $T^{-1}(x,v)$}
    \STATE initialize $x$
    \FOR{$i = 0\ldots n$}
    \STATE sample $v \sim \Normal(v\cond 0, 1)$
    \STATE sample $d \sim \text{Uniform}\{-1,+1\}$
    \STATE propose $[x',v'] = T_{d}(x,v)$, where $T_{d=+1}=T$ and $T_{d=-1}=T^{-1}$
    \STATE evaluate $P = \min\bigg\{1, \frac{p(x',v')}{p(x,v)}\bigg|\frac{\partial T_{d}(x,v)}{\partial [x,v]}\bigg|\bigg\}$
    \STATE accept $x \gets \begin{cases} x', \text{ with probability } P\\
    x, \text{ with probability } (1-P)
    \end{cases}$
    \STATE $x_i \gets x$
    \ENDFOR
    \OUTPUT{samples $\{x_0,\ldots,x_n\}$}
  \end{algorithmic}
  \label{alg:l2hmc}
\end{algorithm}

We recall L2HMC \citep{levy2017generalizing} in Algorithm \ref{alg:l2hmc}.
The core part of the algorithm is the proposal $T(x,v)$, which is learned before the sampling.
The only two differences with A-NICE-MC (see \ref{app:nicemc}) is the form of proposal $T$ (in L2HMC it is not volume-preserving) and the way the proposals are learned.
Since here we do not consider the training stage, we can say that the only difference between A-NICE-MC and L2HMC is the Jacobian of deterministic transformation in the test.
Trick \ref{th:direction} with directional variable $d$ allows for a straightforward formulation of L2HMC in terms of iMCMC.
Consider the joint distribution
\begin{align}
    p(x,v,d) = p(x)\Normal(v\cond 0,1)p(d), \;\;\; p(d) = \text{Uniform}\{-1,+1\},
\end{align}
and the involution
\begin{align}
    f(x,v,d) = [T_{d}(x,v),-d], \;\;\; T_{d=+1}=T,\;\; T_{d=-1}=T^{-1}.
\end{align}
Then it is easy to see that the acceptance probability of iMCMC (Algorithm \ref{alg:imcmc}) is the same as the probability $P$ in Algorithm \ref{alg:l2hmc}.

\subsection{HMC with persistent momentum}
\label{app:horowitz}

\begin{algorithm}[H]
  \caption{HMC with persistent momentum}
  \begin{algorithmic}  
    \INPUT{target density $p(x)$, auxiliary distribution $p(v) = \Normal(v\cond 0,1)$, number of Leap-Frog steps $k$, hyperparameter $\alpha$}
    \STATE initialize $x,v$
    \FOR{$i = 0\ldots n$}
        \STATE update $v \gets v\sqrt{1-\alpha^2} + \alpha\eps, \;\; \eps\sim \Normal(\eps\cond 0,1)$
        \STATE propose $[x',v'] = FL^k(x,v)$
        \STATE evaluate $P = \min\{1, \frac{p(x',v')}{p(x,v)}\}$
        \STATE accept $[x,v] \gets \begin{cases}
        [x',v'], \text{ with probability } P\\
        [x,v], \text{ with probability } (1-P)\\
        \end{cases}$
        \STATE $x_i \gets x$
        \STATE $v \gets -v$
    \ENDFOR
    \OUTPUT{$\{x_0,\ldots, x_n\}$}
  \end{algorithmic} 
  \label{alg:horowitz}
\end{algorithm}

The HMC algorithm with persistent momentum \citep{horowitz1991generalized} is usually formulated as in Algorithm \ref{alg:horowitz}.
The iMCMC formulation of this algorithm can be derived in two ways.
One of the ways is to apply Trick \ref{th:persistent_dir}, we return to it further during the discussion of the generalization of Algorithm \ref{alg:horowitz}.
For illustrative purposes, we firstly describe a straightforward way where we use involution $FL^k$ as a proposal, and compose it with another two iMCMC kernels.
The first kernel  $t_1(x',v',a'\cond x,v,a)$ preserves the joint distribution $p(x,v,a) = p(x)p(v)p(a\cond v)$, where $p(v) = \Normal(v\cond 0,1)$, and $p(a\cond v) = \Normal(a\cond v\sqrt{1-\alpha^2}, \alpha^2)$.
Note that using the involution $f_1(x,v,a) = [x,a,v]$ that just swaps $v$ and $a$ we accepting the new state $[x,a,v]$ with probability $1$.
Indeed,
\begin{align}
    P_1 = \bigg\{1, \frac{p(x)\Normal(a\cond 0,1)\Normal(v\cond a\sqrt{1-\alpha^2}, \alpha^2)}{p(x)\Normal(v\cond 0,1)\Normal(a\cond v\sqrt{1-\alpha^2}, \alpha^2)}\bigg\} = 1.
\end{align}
The second kernel $t_2(x',v'\cond x,v)$ is equivalent to vanilla HMC algorithm with the joint distribution $p(x,v) = p(x) \Normal(v\cond 0,1)$ and the involution $f_2(x,v) = FL^k(x,v)$. The third kernel $t_3(x',v'\cond x,v)$ is equivalent to the flip kernel from Trick \ref{th:persistent_dir}, i.e. iMCMC with the joint distribution $p(x,v) = p(x)\Normal(v\cond 0,1)$ and the involution $f_2(x,v) = [x,-v]$.
Note that the last kernel preserves the distribution without any test since $p(x,v) = p(x,-v)$.

The obtained composition of iMCMC kernels greatly relies on the fact that $p(x,-v) = p(x,v)$, as well as the original proof \citep{horowitz1991generalized}.
However, using the Trick \ref{th:persistent_dir} we can straightforwardly obtain a generalization of this algorithm as depicted in Algorithm \ref{alg:ghorowitz}.
The key idea here is to use an additional directional variable $d \sim \text{Uniform}\{-1,+1\}$ and involution $f(x,v,d) = [T_d(x,v), -d]$, where $T_{d=+1}(x,v) = L^k(x,v)$, and $T_{d=-1}(x,v) = L^{-k}(x,v)$, where $L^{-1}$ is the Leap-Frog inverted in time.
Then we can flip the direction $d$ as in Trick \ref{th:persistent_dir} since $p(d) = p(-d)$.
In the case $p(v) = p(-v)$, and the choice of $t_1(v'\cond v)$ as in Algorithm \ref{alg:horowitz}, we obtain the algorithm equivalent to Algorithm \ref{alg:horowitz}.
Note that in Algorithm \ref{alg:ghorowitz} we consider the case $p(x,v)=p(x)p(v)$ only to be able to apply the explicit version of the Leap-Frog integrator, the same logic applies for implicit integrators as used in RMHMC (Appendix \ref{app:RMHMC}).

\begin{algorithm}[H]
  \caption{Generalized HMC with persistent momentum}
  \begin{algorithmic}  
    \INPUT{target density $p(x)$, auxiliary distribution $p(v)$, number of Leap-Frog steps $k$}
    \INPUT{iMCMC kernel $t_1(v'\cond v)$ for updating $v$}
    \STATE initialize $x,v,d$
    \FOR{$i = 0\ldots n$}
        \STATE update $v \sim t_1(\cdot \cond v)$
        \STATE propose $[x',v',d'] = [T_d(x,v),-d]$, where $T_{d=+1}(x,v) = L^k(x,v)$, and $T_{d=-1}(x,v) = L^{-k}(x,v)$
        \STATE evaluate $P = \min\{1, \frac{p(x',v')}{p(x,v)}\}$
        \STATE accept $[x,v,d] \gets \begin{cases}
        [x',v',d'], \text{ with probability } P\\
        [x,v,d], \text{ with probability } (1-P)\\
        \end{cases}$
        \STATE flip the direction $d \gets -d$
        \STATE $x_i \gets x$
    \ENDFOR
    \OUTPUT{$\{x_0,\ldots, x_n\}$}
  \end{algorithmic} 
  \label{alg:ghorowitz}
\end{algorithm}

\subsection{Gibbs sampling}
\label{app:gibbs}

\begin{algorithm}[H]
  \caption{Gibbs sampling}
  \begin{algorithmic}  
    \INPUT{conditional densities $p(x_k\cond \ldots,x_{k-1},x_{k+1},\ldots)$ of the target distribution $p(x_1,\ldots, x_n)$}
    \STATE initialize $x=(x_1,\ldots,x_n)$
    \FOR{$i = 0\ldots N$}
        \FOR{$k=0\ldots d$}
        \STATE sample $x'_k \sim p(x'_k\cond \ldots,x'_{k-1},x_{k+1},\ldots)$
        \ENDFOR
        \STATE $x[i] \gets (x'_1,\ldots,x'_n)$
        \STATE $x \gets x[i]$
    \ENDFOR
    \OUTPUT{ $\{x[0],\ldots, x[N]\}$}
  \end{algorithmic} 
  \label{alg:gibbs}
\end{algorithm}

Algorithm \ref{alg:gibbs} describes the Gibbs sampling.
Further, we formulate it as the composition of iMCMC kernels, where each kernel is a single step of the inner loop of Algorithm \ref{alg:gibbs}.
That is, for the transition kernel $t_k(x^k\cond x^{k-1})$ we define the joint distribution as
\begin{align}
    p(x_1,\ldots,x_n,v_k) = p(x_1,\ldots,x_n)p(v_k\cond \ldots,x_{k-1},x_{k+1},\ldots),
\end{align}
and the involutive map $f$ as
\begin{align}
    f(x_1,\ldots,x_n,v_k) = [x_1,\ldots,x_{k-1},v_k,x_{k+1},\ldots,x_n,x_k].
\end{align}
It swaps $x_k$ with $v_k$ and leaves the rest of the variables untouched.
The acceptance probability of such a proposal is
\begin{align}
    P = \min\bigg\{1,\frac{p(x_1,\ldots x_{k-1},v_k,x_{k+1},\ldots,x_n)p(x_k\cond \ldots,x_{k-1},x_{k+1},\ldots)}{p(x_1,\ldots,x_n)p(v_k\cond \ldots,x_{k-1},x_{k+1},\ldots)}\bigg\} = 1.
\end{align}
Thus, every proposed point will be accepted and we update variables one by one as in the Gibbs sampling.
The resulted kernel is
\begin{align}
    t(x^n\cond x^0) = \int \prod_{k=1}^{n-1} dx^k \prod_{k=1}^n t_k(x^k\cond x^{k-1}).
\end{align}

Another way to describe the Gibbs sampling is to use Trick \ref{th:persistent_dir}.
Consider the augmented distribution $p(x_1\ldots x_n)p(k)p(d)$, where $p(k) = \text{Uniform}\{1,\ldots,n\}$, and $p(d)=\text{Uniform}\{-1,+1\}$.
Taking the auxiliry distribution as $p(v\cond x_1\ldots x_n, k) = p(v\cond \ldots,x_{k-1},x_{k+1},\ldots)$, we set the involution as
\begin{align}
    f(x_1,\ldots,x_n,v,k,d=+1) = [x_1,\ldots,x_{k-1},v,x_{k+1},\ldots,x_n,x_{k},k+1,-1], \\
    f(x_1,\ldots,x_n,v,k,d=-1) = [x_1,\ldots,x_{k-2},v,x_{k},\ldots,x_n,x_{k-1},k-1,+1],
\end{align}
That is, moving in the positive direction we swap $x_{k}$ and $v$, increment $k \to k+1 \text{ mod } n$ and flip the directional variable $d \to -d$, whereas moving in the negative direction we $x_{k-1}$ and $v$, decrease $k \to k-1 \text{ mod } n$ and also flip the directional variable $d \to -d$.
The acceptance probability of such iMCMC kernel is $1$.
Composing this kernel with the flip of the direction as in Trick \ref{th:persistent_dir}, we obtain a composition of kernels, which every $n$-th sample equals to the samples from Algorithm \ref{alg:gibbs}.

\subsection{Look Ahead HMC}
\label{app:LAHMC}

\begin{algorithm}[H]
  \caption{Look Ahead HMC}
  \begin{algorithmic}  
    \INPUT{target density $p(x)$, auxiliary distribution $p(v) = \Normal(v\cond 0,1)$, hyperparameter $\alpha$}
    \STATE initialize $x,v$
    \FOR{$i = 0\ldots n$}
        \STATE update $v \gets v\sqrt{1-\alpha^2} + \alpha\eps, \;\; \eps\sim \Normal(\eps\cond 0,1)$
        \STATE evaluate $\pi_k = \min\bigg\{1-\sum_{j<k}\pi_j(x,v),\frac{p(FL^k(x,v))}{p(x,v)}\bigg(1-\sum_{j<k}\pi_j(FL^k(x,v))\bigg)\bigg\}$
        \STATE accept $[x,v] \gets \begin{cases}
        L^k(x,v), \text{ with probability } \pi_k(x,v)\\
        [x,-v], \text{ with probability } (1-\sum_k \pi_k(x,v))\\
        \end{cases}$
        \STATE $x_i \gets x$
    \ENDFOR
    \OUTPUT{$\{x_0,\ldots, x_n\}$}
  \end{algorithmic} 
  \label{alg:LAHMC}
\end{algorithm}

The Look Ahead HMC algorithm \citep{sohl2014hamiltonian} operates by proposing several points for acceptance, which are evaluated with different number of steps in the Leap-Frog integrator (see Algorithm \ref{alg:LAHMC}).
The iMCMC formulation of Look Ahead HMC is similar to the formulation of Horowitz's algorithm (see Appendix \ref{app:horowitz}).
The key feature of Look Ahead HMC is that it use a mixture of involutions in the intermediate kernel.

To describe Look Ahead HMC, we use the following composition of iMCMC kernels.
The first kernel  $t_1(x',v',a'\cond x,v,a)$ preserves the joint distribution $p(x,v,a) = p(x)p(v)p(a\cond v)$, where $p(v) = \Normal(v\cond 0,1)$, and $p(a\cond v) = \Normal(a\cond v\sqrt{1-\alpha^2}, \alpha^2)$.
Note that using the involution $f_1(x,v,a) = [x,a,v]$ that just swaps $v$ and $a$ we accepting the new state $[x,a,v]$ with probability $1$.
Indeed,
\begin{align}
    P_1 = \bigg\{1, \frac{p(x)\Normal(a\cond 0,1)\Normal(v\cond a\sqrt{1-\alpha^2}, \alpha^2)}{p(x)\Normal(v\cond 0,1)\Normal(a\cond v\sqrt{1-\alpha^2}, \alpha^2)}\bigg\} = 1.
\end{align}
The second kernel $t_2(x',v',k'\cond x,v,k)$ preserves the joint distribution
\begin{align}
    p(x,v,k) = p(x,v)p(k\cond x,v), \;\;\; p(k\cond x,v) = 1-\sum_{j<k}\pi_j(x,v), \;\;\; k=1,\ldots,K, \;\;\; p(0\cond x,v) = 1-\sum_{k=1}^K \pi_k(x,v) \\
    \pi_k(x,v) = \min\bigg\{1-\sum_{j<k}\pi_j(x,v),\frac{p(FL^k(x,v))}{p(x,v)}\bigg(1-\sum_{j<k}\pi_j(FL^k(x,v))\bigg)\bigg\},
\end{align}
where $p(k\cond x,v)$ defines the index of involution that we apply on the current step.
To be more precise, $k$ defines the number of Leap-Frog steps:
\begin{align}
    f_k(x,v) = FL^k(x,v).
\end{align}
The probability to accept $FL^k(x,v)$ is then
\begin{align}
    P & = \min\bigg\{1,\frac{p(FL^k(x,v))p(k\cond FL^k(x,v))}{p(x,v)p(k\cond x,v)}\bigg\} p(k\cond x,v) = 
    \min\bigg\{p(k\cond x,v),\frac{p(FL^k(x,v))}{p(x,v)}p(k\cond FL^k(x,v))\bigg\} = \\
    & = \min\bigg\{1-\sum_{j<k}\pi_j(x,v),\frac{p(FL^k(x,v))}{p(x,v)}\bigg(1-\sum_{j<k}\pi_j(FL^k(x,v))\bigg)\bigg\} = \pi_k(x,v)
\end{align}
The third kernel $t_3(x',v'\cond x,v)$ simply negates the auxiliary variable $v$.
That is without any resampling, we just apply $f_3(x,v)=[x,-v]$.
Composing all the kernels together we obtain the chain that is equivalent to Algorithm \ref{alg:LAHMC}.

In the formulation above the sign of $v$ plays the role of directional variable $d$ from Trick \ref{th:persistent_dir}.
However, the same can be done explicitly by considering involutions
\begin{align}
    f_k(x,v,d=+1) = [L^k(x,v),-d], \;\;\; f_k(x,v,d=-1) = [FL^kF(x,v),-d]
\end{align}
in the kernel $t_2(x',v',k'\cond x,v,k)$, where $p(d) = \text{Uniform}\{-1,+1\}$.

Further, this Look Ahead technique can be generalized to the case of arbitrary functions $T$ by considering the following family of involutions
\begin{align}
    f_k(x,v,d=+1) = [T^k(x,v),-d], \;\;\; f_k(x,v,d=-1) = [T^{-k}(x,v),-d].
\end{align}

\subsection{Non-Reversible Jump}
\label{app:nrj}

\begin{algorithm}[H]
  \caption{Non-Reversible Jump}
  \begin{algorithmic}
    \INPUT{target density $p(x^{(k)},k)$, auxiliary distributions $q_{k\to k'}(u^{(k)})$, smooth maps $T_{k\to k'}(x^{(k)},u^{(k)})$}
    \STATE initialize $\text{state} = [x^{(k)}, k, \nu]$
    \FOR{$i = 0\ldots n$}
    \STATE $u \sim \text{Uniform}[0,1]$
        \IF{$u \leq \tau$}
        \STATE update $x^{(k)}$ staying in the same model $k$ and fixing the direction $\nu$
        \ELSE
        \STATE unpack $[x^{(k)}, k, \nu] \gets \text{state}$
        \STATE $k' = k + \nu$
        \STATE sample auxiliary $u^{(k)} \sim q_{k\to k'}(u^{(k)})$
        \STATE propose $[x^{(k')}, u^{(k')}] = T_{k\to k'}(x^{(k)}, u^{(k)})$
        \STATE evaluate $P = \min\bigg\{1, \frac{p(x^{(k')},k') q_{k'\to k}(u^{(k')})}{p(x^{(k)},k) q_{k\to k'}(u^{(k)})}\bigg| \frac{\partial T_{k\to k'}}{\partial [x^{(k)}, u^{(k)}]}\bigg|\bigg\}$
        \STATE accept $\text{state} \gets \begin{cases} [x^{(k')}, k', \nu], \text{ with probability } P\\
        [x^{(k)}, k, -\nu], \text{ with probability } (1-P)
        \end{cases}$
        \ENDIF
        \STATE $\text{state}_i \gets \text{state}$
    \ENDFOR
    \OUTPUT{samples $\{\text{state}_0,\ldots,\text{state}_n\}$}
  \end{algorithmic}
  \label{alg:nrj}
\end{algorithm}

We provide the pseudo-code for Non-Reversible Jump scheme \citep{gagnon2019non} in Algorithm \ref{alg:nrj}.
Further, we describe this algorithm in terms of iMCMC using Trick \ref{th:persistent_dir}.
To build the first kernel $t_1(\cdot \cond \cdot)$, we consider the following joint distribution
\begin{align}
    p(x^{(k)},u^{(k)},v^{(k)},k,\nu,m) = p(x^{(k)},k)p(\nu)p(u^{(k)}\cond k,\nu)p(m)p(v^{(k)}\cond k),
\end{align}
where $p(\nu) = \text{Uniform}\{-1,+1\}$ is analogue of direction $d$ in Trick \ref{th:direction}; $p(m) = \text{Bernoulli}(\tau, 1-\tau)$ defines the index of involution applied; $p(v^{(k)}\cond k)$ and $p(u^{(k)}\cond k,\nu)$ define auxiliary variables, which we choose as $p(u^{(k)}\cond k,\nu) = q_{k\to k+\nu}(u^{(k)})$ and $p(v^{(k)}) = q_{k\to k}(v^{(k)})$.
With probability $1-\tau$ (when $m=1$), we apply involution
\begin{align}
    f_1(x^{(k)},u^{(k)}, v^{(k)}, k, \nu) = [T_{k\to(k+\nu)}(x^{(k)},u^{(k)}), v^{(k)}, k+\nu, -\nu, v^{(k)}] = [x^{(k+\nu)},u^{(k+\nu)}, v^{(k)}, k+\nu, -\nu], \\ 
    T_{k'\to k}(x^{(k')},u^{(k')}) = T_{k\to k'}^{-1}(x^{(k')},u^{(k')}) = [x^{(k)},u^{(k)}].
\end{align}
That is, based on indices $k$ and $k+\nu$ we choose a smooth map that we apply to $x^{(k')},u^{(k')}$; we also update $k \to k+\nu$ and negate the direction $\nu$.
The acceptance probability for such a proposal is
\begin{align}
    P = \min \bigg\{1, \frac{p(x^{(k+\nu)},k+\nu)p(u^{(k+\nu)}\cond k+\nu,-\nu)}{p(x^{(k)},k)p(u^{(k)}\cond k,\nu)}\bigg| \frac{\partial T_{k\to (k+\nu)}}{\partial [x^{(k)}, u^{(k)}]}\bigg|\bigg\},
\end{align}
which is equivalent to the acceptance probability in Algorithm \ref{alg:nrj}, when we denote $k'=k+\nu$ and $p(u^{(k)}\cond k,\nu) = q_{k\to k+\nu}(u^{(k)})$.
With probability $\tau$ (when $m=0$), we apply involution
\begin{align}
    f_0(x^{(k)}, v^{(k)},u^{(k)}, k, \nu) = [T_{k\to k}(x^{(k)},v^{(k)}),u^{(k)}, k, \nu], \;\;\; T_{k\to k}(x^{(k)},v^{(k)}) = T_{k\to k}^{-1}(x^{(k)},v^{(k)}),
\end{align}
which does not change neither $k$ nor $\nu$.
Here we also apply involutive smooth map $T_{k\to k}$ to the vector $[x^{(k)},v^{(k)}]$ instead of $[x^{(k)},u^{(k)}]$.
Without the loss of generality, we can treat the case of $m=0$ to be equivalent to the corresponding update when $u\leq \tau$ in Algorithm \ref{alg:nrj}.

As well as in Trick \ref{th:persistent_dir}, we combine the obtained kernel $t_1$ on the joint distribution $p(x^{(k)},u^{(k)},v^{(k)},k,\nu,m)$ with the kernel $t_2$ on the same distribution.
Applying $t_2$ we do not resample any variables, instead we use the following involution
\begin{align}
    f(\nu,m=0) = [\nu,m], \;\;\; f(\nu,m=1) = [-\nu,m].
\end{align}
The rest of the variables remains the same.
Based on the value of $m$ we change only $\nu$ to obtain the persistent irreversible movement in the case when $\nu$ was negated by the kernel $t_1$.
The combination of kernels $t_1$ and $t_2$ yields the sampler that is equivalent to Non-Reversible Jump scheme (Algorithm \ref{alg:nrj}).

\subsection{Lifted Metropolis-Hastings}
\label{app:lmh}

Firstly, we recall a general approach of Lifting in \citep{turitsyn2011irreversible} following the formulation from \citep{bierkens2017piecewise}.
Lifting modifies the reversible kernel $T$ on the state space $X$ by splitting each state $x \in X$ in two replicas: $\{x,+\}$ and $\{x,-\}$.
Then, for each replica, the authors introduce its own transition kernel: $T^{(+)}$ for positive replicas and $T^{(-)}$ for negative ones.
These transition kernels must satisfy
\begin{align}
    T(x,y)^{(+)}p(x) = T(y,x)^{(-)}p(y), \;\;\; \forall x\neq y,
\end{align}
where $p$ is the target distribution.
The kernels $T^{(+)}$ and $T^{(-)}$ define in-replica transitions and are obtained from the original kernel $T$ by splitting the support of $T$ using some decision function $\eta: X \to \mathbb{R}$.
For non-diagonal elements $x\neq y$ these transitions can be written as
\begin{align}
    T^{(+)}(x,y) = \begin{cases}
    T(x,y), \text{ if } \eta(y) \geq \eta(x),\\
    0, \;\;\;\;\;\;\;\;\; \text{ if } \eta(y) < \eta(x)
    \end{cases} \text{ and } \;\;\;
    T^{(-)}(x,y) = \begin{cases}
    0, \;\;\;\;\;\;\;\;\; \text{ if } \eta(y) > \eta(x),\\
    T(x,y), \text{ if } \eta(y) \leq \eta(x)
    \end{cases}.
\end{align}
Inter-replica transitions are defined as
\begin{align}
    T^{(-,+)}(x) = \max\bigg\{0, \sum_{y: y\neq x} T^{(+)}(x,y) - T^{(-)}(x,y)\bigg\}, \\
    T^{(+,-)}(x) = \max\bigg\{0, \sum_{y: y\neq x} T^{(-)}(x,y) - T^{(+)}(x,y)\bigg\}.
\end{align}
Where $T^{(+,-)}$ define the transition probability from positive replicas to negative ones.
Finally, the diagonal elements of $T^{(+)}$ and $T^{(-)}$ are defined as follows.
\begin{align}
    T^{(+)}(x,x) = 1 - T^{(+,-)}(x) - \sum_{y: y\neq x} T^{(+)}(x,y), \;\;\; T^{(-)}(x,x) = 1 - T^{(-,+)}(x) - \sum_{y: y\neq x} T^{(-)}(x,y)
\end{align}
Note that 
\begin{align}
    T^{(+)}(x,x) = T^{(-)}(x,x) = \min \bigg\{1-\sum_{y: y\neq x} T^{(-)}(x,y),1-\sum_{y: y\neq x} T^{(+)}(x,y)\bigg\}.
\end{align}
The whole transition kernel on the extended space is defined as
\begin{align}
    \mathcal{T} = \begin{bmatrix}
    T^{(+)} & T^{(+,-)} \\
    T^{(-,+)} & T^{(-)} \\
    \end{bmatrix}.
\end{align}

To describe Lifting in terms of iMCMC we follow Trick \ref{th:lifting} introducing the directional variable $p(d) = \text{Uniform}\{-1,+1\}$, which define the proposal we are currently using to sample new state.
Further, we compose this kernel with the flip of $d$ to obtain an irreversible kernel.
That is, the first kernel $t_1$ operates on the following distribution.
\begin{gather}
    p(x,v,d) = p(x)p(d)q(v\cond x, d), \\ 
    q(v\cond x,+1) = T^{(+)}(x,v) \;\; \forall v\neq x, \;\;\; q(v\cond x,-1) = T^{(-)}(x,v) \;\; \forall v\neq x, \\
    q(x\cond x,+1) = 1-\sum_{v: v\neq x}T^{(+)}(x,v), \;\;\; q(x\cond x,-1) = 1-\sum_{v: v\neq x}T^{(-)}(x,v)
\end{gather}
The involutive map is then
\begin{gather}
    f_1(x,v,d) = [v,x,-d],
\end{gather}
which is just the swap of $x$ and $v$ and the negation of $d$.
Kernel $t_1$ is then obtained by substitution of $p(x,v,d)$ and $f_1(x,v,d)$ into Algorithm \ref{alg:imcmc}.
Then we compose the first kernel $t_1$ with the kernel $t_2$ that just negate the directional variable one more time applying the involution $f_2(x,v,d) = [x,v,-d]$.
The composition of kernels $t_1$ and $t_2$ we denote as $t(x',v',d'\cond x,v,d)$.

To prove that the iMCMC formulation is equivalent to the original chain we consider three following cases.
The first case is the transition to the new state $v \neq x$ staying in the same replica (same direction $d$).
\begin{align}
    \forall x\neq v, \;\; t(v,+1\cond x,+1) = q(v\cond x, +1)\min\bigg\{1,\frac{p(v)q(x\cond v, -1)}{p(x)q(v\cond x, +1)}\bigg\} = T^{(+)}(x,v)\min\bigg\{1,\frac{p(v)T^{(-)}(v,x)}{p(x)T^{(+)}(x,v)}\bigg\} = T^{(+)}(x,v).
\end{align}
Note that the directional variable remains the same because of the double negation: firstly in $f_1$ and then in $f_2$.
The second case is the staying in the same state $x$ with the same direction.
\begin{align}
    t(x,+1\cond x,+1) &= q(x\cond x, +1)\min\bigg\{1,\frac{p(x)q(x\cond x, -1)}{p(x)q(x\cond x, +1)}\bigg\} \\ 
    &= (1-\sum_{v: v\neq x}T^{(+)}(x,v))\min\bigg\{1,\frac{p(x)(1-\sum_{v: v\neq x}T^{(-)}(x,v))}{p(x)(1-\sum_{v: v\neq x}T^{(+)}(x,v))}\bigg\} = T^{(+)}(x,x).
\end{align}
The last case is the inter-replica transition of Lifting, which corresponds to the rejection in its iMCMC formulation.
\begin{align}
    t(x,-1\cond x,+1) & = 1 - \sum_{v}t(v,+1\cond x,+1) = 1-\sum_{v: v\neq x}t(v,+1\cond x,+1) - t(x,+1\cond x,+1) = \\ 
    & = 1-\sum_{v: v\neq x}T^{(+)}(x,v) - T^{(+)}(x,x) = \\ 
    & = 1-\sum_{v: v\neq x}T^{(+)}(x,v) - \min \bigg\{1-\sum_{v: v\neq x} T^{(-)}(x,v),1-\sum_{v: v\neq x} T^{(+)}(x,v)\bigg\} = \\
    & = \max\bigg\{0, \sum_{v: v\neq x} T^{(-)}(x,v) - T^{(+)}(x,v)\bigg\} = T^{(+,-)}(x).
\end{align}

\section{Experiments}

\subsection{Distributions}
\label{app:dists}

Here we provide analytical forms of considered target distributions.
Target density for MoG2 is:
\begin{equation}
p(x) = \frac{1}{2}\Normal(x | \mu_1, \sigma_1) + \frac{1}{2}\Normal(x | \mu_2, \sigma_2)
\end{equation}
where $\mu_1 = [2, 0]$, $\mu_2 = [-2, 0]$, $\sigma_1^2 = \sigma_2^2 = \begin{bmatrix}
     0.5 & 0 \\
     0 & 0.5
\end{bmatrix}$. 

For the Bayesian logistic regression, we define likelihood and prior as
\begin{equation}
    p(y=1\cond x, \theta) = \frac{1}{1+\exp(-x^T\theta_w+\theta_b)}, \;\;\; p(\theta) = \Normal(\theta\cond 0, 0.1).
\end{equation}
Then the unnormalized density of the posterior distribution for a dataset $D=\{(x_i,y_i)\}_i$ is
\begin{equation}
    p(\theta\cond D) \propto \prod_i p(y_i\cond x_i, \theta) p(\theta).
\end{equation}
We sample from the posterior distribution on three datasets: German ($25$ covariates, $1000$ data points), Heart ($14$ covariates, $532$ data points) and Australian ($15$ covariates, $690$ data points).
We provide all the data with the code in supplementary.

\subsection{Effective sample size}
\label{app:ess}

The effective sample size (ESS) is defined as the reciprocal of the autocorrelation time.
It is designed to represent the number of truly independent samples that would be equivalent to a correlated sample drawn using the chain. 
There are several approaches to evaluation of autocorrelation time \citep{thompson2010comparison}.
One of the most common approaches is the initial sequence estimators.
That is, the autocorrelation $\rho$ of sequence $\{X_i\}_{i=1}^n$ is estimated as
\begin{align}
    \widehat{\rho} = 1 + 2\sum_{k=1}^{\infty}\rho_k, \;\;\; \widehat{\rho}_k = \frac{1}{ns^2}\sum_{i=1}^{n-k}(X_i - \overline{X}_n)(X_{i+k} - \overline{X}_n),
\end{align}
where $\overline{X}_n$ and $s^2$ are the sample mean and variance of the sequence.
Further, assuming the reversibility of the chain, the consecutive pair $\rho_i + \rho_{i+1}$ is always positive \citep{geyer1992practical}.
Thus, one can obtain initial positive sequence estimator by truncating the negative values of the sums $\widehat{\rho}_i + \widehat{\rho}_{i+1}$.

However, the initial positive sequence estimator fails to converge to the true autocorrelation in some cases \citep{thompson2010comparison}.
Moreover, in this paper we cannot rely on the reversibility of the chain since we compare reversible chains with their irreversible analogues.
That is why we turn to the batch-means estimator of the autocorrelation time, which operates as follows.
It divides the initial sequence $\{X_i\}_{i=1}^n$ into subsequences (batches) of size $m$ and evaluate sample means of each batch.
Then we estimate $\rho$ as
\begin{align}
    \widehat{\rho} = m \frac{s^2_m}{s^2},
\end{align}
where $s^2_m$ is the sample variance of batch means.
For the choice of $m$ we follow \citep{thompson2010comparison}, and take $n^{1/3}$ batches of the size $m=n^{2/3}$.
For multivariate distributions we follow the common practice of evaluating the minimal ESS across all dimensions.

To include computation efforts into the performance evaluation, we calculate ESS per second.
We run all the algorithms on a single GPU with batch size $100$ sampling $20000$ samples, and discarding first $1000$ for burn-in.
The final formula is 
\begin{align}
    \text{ESS/s} = \frac{1}{\rho}\frac{\text{number of samples}}{\text{run time}}.
\end{align}

\subsection{Irr-MALA}
\label{app:neg-mala}

Following Trick \ref{th:persistent_dir}, we modify the original algorithm by introducing the directional variable $p(d) = \text{Uniform}\{-1,+1\}$.
For the first kernel $t_1(x',v',d'\cond x,v,d)$, the joint distribution is
\begin{align*}
    p(x,v,d) = p(x)\Normal(v\cond x + d\eps \nabla_x\log p(x), 2\eps)p(d),
\end{align*} 
and the involutive map is 
\begin{align*}
    f_1(x,v,d) = [v,x,-d\cdot\text{sign}(\nabla_x\log p(x)^T\nabla_v\log p(v))].
\end{align*}
Then the acceptance probability is
\begin{align}
    P = \min\bigg\{1, \frac{p(v)\Normal(x\cond v + d'\eps \nabla_v\log p(v), 2\eps)}{p(x)\Normal(v\cond x + d\eps \nabla_x\log p(x), 2\eps)}\bigg\}, \;\;\; d' = -d\cdot\text{sign}\bigg(\nabla_x\log p(x)^T\nabla_v\log p(v)\bigg).
\end{align}
Note that defining the sign of the gradient $\nabla_v\log p(v)$ via $d'$, we ensure that the mean $v + d'\eps \nabla_v\log p(v)$ will be close to the initial point $x$.
The second kernel $t_2(x',v',d'\cond x,v,d)$, as well as in Trick \ref{th:persistent_dir}, is just the flip of the direction $d$.
That is, we do not resample any variables, instead we apply the involution $f_2(x,v,d) = [x,v,-d]$.
Combining the kernels $t_1$ and $t_2$, we obtain an irreversible chain.
We provide the pseudo-code in Algorithm \ref{alg:irr-mala}.

\begin{algorithm}[H]
  \caption{Irr-MALA}
  \begin{algorithmic}
    \INPUT{target density $p(x)$, step size $\eps$}
    \STATE initialize $[x,d]$
    \FOR{$i = 0\ldots n$}
    \STATE sample $v \sim \Normal(v\cond x + d\eps \nabla_x\log p(x), 2\eps)$
    \STATE evaluate $d' = -d\cdot\text{sign}\bigg(\nabla_x\log p(x)^T\nabla_v\log p(v)\bigg)$
    \STATE evaluate $P = \min\bigg\{1, \frac{p(v)\Normal(x\cond v + d'\eps \nabla_v\log p(v), 2\eps)}{p(x)\Normal(v\cond x + d\eps \nabla_x\log p(x), 2\eps)}\bigg\}$
    \STATE accept $[x,d] \gets \begin{cases} [v,d'], \text{ with probability } P\\
    [x,d], \text{ with probability } (1-P)
    \end{cases}$
    \STATE $d \gets -d$
    \STATE $x_i \gets x$
    \ENDFOR
    \OUTPUT{samples $\{x_0,\ldots,x_n\}$}
  \end{algorithmic}
  \label{alg:irr-mala}
\end{algorithm}

\subsection{Irr-NICE-MC}
\label{app:irr-nice}

The irreversible analog of A-NICE-MC \citep{song2017nice} is easily obtained from the original algorithm (see \ref{app:nicemc}) by composing it with two additional kernels.
The first kernel $t_1(x',v',d',a'\cond x,v,d,a)$ operates by changing only the auxiliary variable $v$.
That is, consider the joint distribution 
\begin{align}
    p(x,v,d,a) = p(x)p(v)p(d)p(a\cond v), \;\; p(v)=\Normal(v\cond 0,1), \;\; p(a\cond v) = \Normal(a\cond v\sqrt{1-\alpha^2},\alpha^2), \;\; p(d)=\text{Uniform}\{-1,+1\}.
\end{align}
And the involution $f_1(x,v,d,a) = [x,a,d,v]$ that just swap $a$ and $v$.
Note that the acceptance probability
\begin{align}
    P_1 = \bigg\{1, \frac{p(x)\Normal(a\cond 0,1)\Normal(v\cond a\sqrt{1-\alpha^2}, \alpha^2)}{p(x)\Normal(v\cond 0,1)\Normal(a\cond v\sqrt{1-\alpha^2}, \alpha^2)}\bigg\} = 1.
\end{align}
The second kernel $t_2(x',v',d'\cond x,v,d)$ is equivalent to the A-NICE-MC kernel with only difference that we do not resample $d$ at each step.
The joint distribution of this kernel is
\begin{align}
    p(x,v,d) = p(x)p(v)p(d), \;\;\; p(v)=\Normal(v\cond 0,1), \;\;\; p(d)=\text{Uniform}\{-1,+1\}.
\end{align}
And the involutive map is
\begin{align}
    f_2(x,v,d) = [T_d(x,v),-d], \;\;\; T_{d=+1} = T, \;\; T_{d=-1}=T^{-1}.
\end{align}
The last kernel $t_3(x',v',d'\cond x,v,d)$ operates on the same joint distribution $p(x,v,d)$, and just negate the directional variable $d$ with involution $f_3(x,v,d) = [x,v,-d]$.
Combining all three kernels, we obtain irreversible modification of A-NICE-MC.
See pseudo-code in Algorithm \ref{alg:irr-nice}.

\begin{algorithm}[H]
  \caption{Irr-NICE-MC}
  \begin{algorithmic}
    \INPUT{target density $p(x,v) = p(x)\Normal(v\cond 0,1)$, NICE-proposal $T(x,v)$ and $T^{-1}(x,v)$}
    \STATE initialize $[x,v,d]$
    \FOR{$i = 0\ldots n$}
    \STATE sample $\widehat{v} \sim \Normal(\widehat{v}\cond v\sqrt{1-\alpha^2}, \alpha^2)$
    \STATE propose $[x',v'] = T_{d}(x,\widehat{v})$, where $T_{d=+1}=T$ and $T_{d=-1}=T^{-1}$
    \STATE evaluate $P = \min\bigg\{1, \frac{p(x',v')}{p(x,v)}\bigg\}$
    \STATE accept $[x,v,d] \gets \begin{cases} [x',v',-d], \text{ with probability } P\\
    [x,\widehat{v},d], \text{ with probability } (1-P)
    \end{cases}$
    \STATE $d \gets -d$
    \STATE $x_i \gets x$
    \ENDFOR
    \OUTPUT{samples $\{x_0,\ldots,x_n\}$}
  \end{algorithmic}
  \label{alg:irr-nice}
\end{algorithm}

\end{document}